\documentclass{article}

\usepackage[main,preprint]{neurips_2026}
\usepackage{amsthm}
\usepackage{amsmath}
\usepackage{graphicx}
\usepackage{subcaption}
\theoremstyle{plain}
\newtheorem{theorem}{Theorem}[section]

\newtheorem{lemma}[theorem]{Lemma}

\theoremstyle{definition}
\newtheorem{definition}[theorem]{Definition}

\theoremstyle{remark}
\newtheorem{remark}[theorem]{Remark}

\usepackage[utf8]{inputenc} % allow utf-8 input
\usepackage[T1]{fontenc}    % use 8-bit T1 fonts
\usepackage{hyperref}       % hyperlinks
\usepackage{url}            % simple URL typesetting
\usepackage{booktabs}       % professional-quality tables
\usepackage{amsfonts}       % blackboard math symbols
\usepackage{nicefrac}       % compact symbols for 1/2, etc.
\usepackage{microtype}      % microtypography
\usepackage{xcolor}         % colors

\title{In-Context Multi-Operator Learning with DeepOSets}

\renewcommand{\citet}{\cite}
\author{%
  Shao-Ting Chiu\thanks{Equal contribution.} \\
  Department of Electrical\\ and Computer Engineering\\
  Texas A\&M University\\
  College Station, TX 77843 \\
  \texttt{stchiu@tamu.edu} \\
  \And
  Aditya Nambiar\footnotemark[1] \\
  Department of Mathematics\\
  Texas A\&M University\\
  College Station, TX 77843 \\
  \texttt{aditnambiar@tamu.edu} \\
  \And
  Ali Syed \\
  Department of Mathematics\\
  Texas A\&M University\\
  College Station, TX 77843 \\
  \texttt{saabidi1@tamu.edu} \\
  \And
  Jonathan W. Siegel \\
  Department of Mathematics\\
  Texas A\&M University\\
  College Station, TX 77843 \\
  \texttt{jwsiegel@tamu.edu} \\
  \And
  Ulisses Braga-Neto \\
  Department of Electrical\\ and Computer Engineering\\
  Texas A\&M University\\
  College Station, TX 77843 \\
  \texttt{ulisses@tamu.edu} \\
}

\begin{document}

\maketitle

\begin{abstract}
  An important application of neural networks in scientific computing is non-linear operator learning. In this framework, a neural network is trained to fit a non-linear map between infinite dimensional spaces, for example, the solution operator of ordinary and partial differential equations. Recently, inspired by the discovery of in-context learning for large language models, an even more ambitious paradigm has been explored, called multi-operator learning. In this approach, a neural network is trained to learn many different operators at the same time. At inference time, the network is passed a small number of example input and output pairs to disambiguate the desired operator. In this work, we provide a precise mathematical formulation of the multi-operator learning problem, propose a parameter-efficient, attention-free architecture, called DeepOSets, and prove its universality in the in-context multi-operator learning task. We perform experiments demonstrating the efficacy of DeepOSets on several benchmarks, including Burgers' equation, 1D and 2D Darcy flow, and a variety of initial-value and boundary-value differential equations. In a comparison with the transformer-based ICON model on the 1D Darcy problem, under a small training budget (10k steps), DeepOSets attained lower error at every context size with 19\% of the parameters, while with an extended budget (100k steps), ICON attains 26--41\% lower error, but at the cost of roughly 218$\times$ more training time, 12$\times$ higher inference latency, and 5.3$\times$ more parameters.
\end{abstract}

\section{Introduction}

Learning non-linear operators from data using neural networks has become a foundational application of deep learning in scientific computing \citep{ludeeponet2021,li2020fourier}. In learning the solution operator of a parametric PDE from sample data, one is not interested in merely obtaining a single solution for a single PDE, but instead learning the mapping between parameter and solution, so that new solutions corresponding to new parameter values can be obtained quickly and without further training. Furthermore, both the parameter and the solution are typically functions in infinite-dimensional spaces. The trained operator allows quick execution in many-query problems, such as design optimization, uncertainty quantification, or real-time processing, which is infeasible to do with expensive high-fidelity solvers. 
%The most well-known neural operators are DeepONets \citep{ludeeponet2021} and Fourier Neural Operators \citep{li2020fourier}, which are both based on specific deep neural network architectures.

On the other hand, many problems of interest in science and engineering involve solving complex PDEs with various unknown parameters or terms, in which case one has  
multiple PDE solution operators, and the neural network model needs to be trained on data from a mixture of PDEs. This is a multi-operator learning problem, also known as a PDE foundation model\citep{sun2025towards,herde2024poseidon,zhang2025probabilistic_operator_learning}. To disambiguate the correct operator during inference time, one needs additional data, which could be used to fine-tune the neural network, or could be presented in a prompt, in which case there are no weight updates. The latter approach is known as in-context learning (ICL). 
At this time, In-Context Operator Networks (ICON) \citep{yang2023context}, and its subsequent variants \citep{yang2024pde,yang2023ICON-LM,zhang2025GenICON}, is the best-known approach to in-context learning of PDE solution operators. ICON is based on a GPT-style architecture and thus has quadratic complexity in prompt length, due to the dot-product self-attention mechanism. This can introduce significant computational overhead and loss of accuracy with very long prompts. 

A theoretical study of in-context multi-operator learning remains incomplete. In particular, it is not known whether existing architectures such as ICON are universal approximators. One key difficulty is that the lack of a precise mathematical formulation of the multi-operator learning problem. In this paper, we provide such a mathematical formulation, by framing the problem as learning a \textit{compact set of operators} between Banach spaces.
Furthermore, we propose an alternative approach to in-context learning of PDE solution operators, which processes the prompt non-autoregressively and without using self-attention, being therefore much faster during both training and inference time. This approach is based on DeepOSets, a neural architecture that combines set learning via the DeepSets architecture \citep{zaheer2017deep} with the DeepONets neural operator architecture \cite{ludeeponet2021}. In DeepOSets, the DeepSets layer acts as an encoder of the prompt into a latent representation that is the input, along with the query, to the branch path of a DeepONet. The DeepSets layer confers linear complexity in the length of the prompt, in contrast to the quadratic complexity of self-attention, and a built-in permutation-invariance inductive bias to the order of the examples in the prompt.

% DeepOSets has been shown previously to display ICL capabilities in regression problems~\citep{chiu2024deeposets}. Here, we introduce architectural modifications to DeepOSets that allow it to act as a multi-operator learner for parametric differential equations. This is accomplished by discretizing a variable number of solution and parameter function pairs, which are entered into the prompt and encoded by the DeepSets layer into latent representation. The latter is concatenated to the discretization of the query parameter function and entered directly into the branch path of the DeepONet. A detailed description of our architecture can be found in Section \ref{model-section} and Figure \ref{fig:model}.

Crucially, we prove that the DeepOSets architecture is universal for the in-context multi-operator learning problem. Specifically, we prove that it can approximate {\em uniformly} any compact set of continuous solution operators. This means that DeepOSets are not only universal approximators of single solution operators, like DeepONet, but they are universal approximators of multiple operators, in the sense that a single DeepOSets architecture exists that can approximate {\em any} operator in the class to the same fixed degree of accuracy. This guarantees that DeepOSets can learn in-context the solution operators of multiple differential equations.  

Experiments with the Burgers' equation, 1D and 2D Darcy flow, and a variety of initial-value and boundary-value differential equations demonstrate that DeepOSets can accurately predict the solution to a new differential equation not seen during training. 
Furthermore, we conducted a direct controlled comparison with ICON on the 1D Darcy flow problem considered previously, which showed that under a small training budget DeepOSets outperformed ICON at every context size while using 19\% of the parameters, while at large number of training steps, ICON is more accurate, with a relative error reduction from 26\% to 41\%, but at 5.2 times more parameters and roughly 218 times more wall-clock training time than DeepOSets (8.74 hours versus 11.6 seconds).

\subsection{Our Contributions}
We precisely formulate the multi-operator learning problem, propose the DeepOSets architecture for solving this problem, and prove its universality. We also provide experiments demonstrating the efficacy of our approach.
\begin{enumerate}
    \item We provide a \textit{precise mathematical formulation} of the in-context multi-operator learning problem and define universal approximation for this problem within Section \ref{sec:operator-framework} and Definition \ref{universality-definition} respectively.
    \item We prove that the DeepOSets architecture defined in Section \ref{model-section} is \textit{universal} for in-context multi-operator learning within Section \ref{sec:universality for in-context} in Theorem \ref{main-universality-theorem}. This is to the best of our knowledge the first universal approximation result for in-context multi-operator learning.
    \item {\color{black} We demonstrate experimentally that DeepOSets is capable of learning compact classes of operators, including the operators corresponding to Burgers' equation and Darcy flow with variable diffusion, in the Experimental Results (Section \ref{sec:experiments}).}
\end{enumerate}

\section{Related Work}
The universal approximation theorem for multi-layer perceptrons, which forms the basis for more sophisticated universal approximation theorems, was proved in \citet{cybenko1989approximation,hornik1991approximation}. Universal approximation of operators using neural networks was first proved in \citet{chen1995universal}, following previous results from \citet{chen1993approximations,chen1995approximation}. This result was used to design the DeepONet operator learning architecture \cite{ludeeponet2021}. Subsequently, various universal architectures for operator learning have been proposed, a notable example being Fourier Neural Operators \cite{li2020fourier}. These architectures typically evaluate the input function on a \textit{fixed} grid, and an architecture compatible with variable input grids was proposed and analyzed in \citet{prasthofer2022variable}.  

MIONet is to the best of our knowledge the first approach to attempt multi-input operator learning, by encoding several input functions via multiple branch network (and a shared trunk network) of a DeepONet, though it still learns a single operator \cite{jin2022mionet}. POSEIDON is a large foundation model for PDEs based on a scalable Operator Transformer (scOT) with a vision transformer architecture, is pretrained on compressible Euler and incompressible Navier-Stokes equations and applied to 15 downstream tasks, showing better sample efficiency and accuracy than baselines \cite{herde2024poseidon}. PROSE-PDE is a transformer-based multi-operator learning approach trained on one-dimensional time-dependent nonlinear PDEs that encodes  both numerical and symbolic data types for forward and inverse PDE problems \cite{sun2025towards}. LeMON is based on pretraining and fine-tuning for multi-operator learning, and introduces low-rank adaptation to reduce computational cost \cite{sun2024lemon}. MT-DeepONet introduces modifications to the branch network of a DeepONet to accommodate various functional forms of parameterized coefficients and multiple geometries \cite{kumar2025synergistic}.  

In-context learning (ICL) is a phenomenon first observed in large language models \cite{NEURIPS2020_1457c0d6,NEURIPS2021_01b7575c}, whereby these models are able to learn new tasks without fine-tuning from examples presented to them in the prompt. This has sparked substantial theoretical research effort into the mechanism underlying this phenomenon. Transformers have been theoretically shown to exhibit ICL on a variety of linear models, including least squares, ridge regression, Lasso, generalized linear models and linear inverse problems \cite{garg2022can,bai2023transformers,zhang2024trained,von2023transformers}. Generalization error for transformer-based ICL for data lying on a low-dimensional manifold has been studied in \cite{shen2026understanding}. In-context learning for differential equation solution operators was introduced by ICON \cite{yang2023context}, with a follow-up study on 1D scalar conservation laws providing a detailed methodology for applying ICON to time-evolution PDEs \cite{yang2024pde}. ICON-LM integrates ICON concepts with language-model architectures \cite{yang2023ICON-LM}. GenICON introduces a generative, probabilistic approach to ICON \cite{zhang2025GenICON}. In-context learning of linear PDEs using transformers has also been studied in \cite{cole2024provable}.

\section{Mathematical Formulation of In-context Multi-Operator Learning}\label{sec:operator-framework}
In-context operator learning involves a neural network learning a class of operators, and identifying a specific operator from data, called in-context examples. In this section, we give a precise mathematical formulation of this inverse problem. Based upon this, we will be able to give a precise definition of universality for this problem.

Let $K_1, K_2\subset \mathbb{R}^d$ be compact subsets of some Euclidean space. We will consider continuous maps between the Banach spaces $X := C(K_1)$ and $Y := C(K_2)$, whose domain is a compact subset $V\subset X$. Note that the dimensionality of $K_1$ and $K_2$ need not be the same in general, but to simplify the presentation we restrict to this case in the following. The minor adjustments required to handle the more general case where $K_1\subset \mathbb{R}^{d_1}$ and $K_2\subset \mathbb{R}^{d_2}$ are left to the reader.

In traditional operator learning, a continuous function $G:V\rightarrow Y$ is approximated using a neural network. The input function $u\in V$ is discretized at a fixed set of points $x_1,...,x_k\in K_1$, and the values $u(x_1),...,u(x_k)$ together with an evaluation point $y\in K_2$ are passed to a neural network whose output approximates $G(u)(y)$. Universality in this context means that for any $G$ and $\varepsilon > 0$, there exists a grid $\textbf{x} := \{x_1,...,x_k\}\in K_1$ (depending on $G$ and $\varepsilon$) and a neural network $G_\varepsilon$ such that
\begin{align*}
    \left|G_\varepsilon(u(\textbf{x}),y) - G(u)(y)\right| < \varepsilon
\end{align*}
for any $u\in V$ and $y\in K_2$ \cite{chen1995universal,ludeeponet2021}. Here we denote by $$u(\textbf{x}) := \left(u(x_i)\right)_{i=1}^k\in \mathbb{R}^k$$
the function evaluated at the grid $\textbf{x}$. 

Instead of learning a single operator $G:V\rightarrow Y$, in-context multi-operator learning considers an entire class of operators. To formulate this, we consider the space of continuous operators:
$$\mathcal{O} := \{G:V\rightarrow Y~\text{continuous}\} $$
equipped with the induced operator norm given by
$$
\|G\|_{\mathcal{O}}:=\sup_{x \in V}{\|G(x)\|_{Y}}.
$$
We let $\mathcal{K} \subset \mathcal{O}$ be a compact subset of this space of operators. $\mathcal{K}$ is the class of operators that we wish to learn. In order to disambiguate the desired operator $G\in \mathcal{K}$, we pass the network a sequence of \textit{in-context examples} $(u_1,G(u_1)),...,(u_m,G(u_m))$ in addition to the input function $u\in K$ and the evaluation point $y\in K_2$. All of the functions $u_i$, $u$, and $G(u_i)$ are discretized on fixed grids $\textbf{x} = (x_1,...,x_k)$ and $\textbf{y} = (y_1,...,y_n)$ in $K_1$ and $K_2$, respectively. We remark that one could additionally consider variable grids in $K_1$ and $K_2$, but we leave this to future work.

\subsection{Examples of in-context operator learning}
Let us begin by giving some examples of compact sets of operators and instances of in-context operator learning.

\textbf{Finite sets of operators.} Any finite set of operators $\mathcal{K} = \{G_1,...,G_p\}$ is a compact set. In this case, in-context operator learning would identify and evaluate the correct operator $G = G_q$ for some $q$ from the in-context examples. This is rather trivial example of in-context multi-operator learning.

\textbf{Continuously parameterized sets of operators.} If the set of operators is continuously parameterized by a compact set, i.e., $\mathcal{K} = \{G_\mu:~\mu\in K'\}$ where $K'$ is a compact set (typically in $\mathbb{R}^d$ or in a Banach space), then $\mathcal{K}$ is compact. This is the typical application that we have in mind. As an example, consider the Darcy flow with variable permeability:
\begin{equation}
    -\nabla\cdot(k(x)\nabla u(x)) = f(x)~\text{in $\Omega$},~~u(x) = 0~\text{on $\partial \Omega$}.
\end{equation}
The solution operator mapping $f\rightarrow u$ is parameterized by the permeability $k$, and it is known that this parameterization is continuous with respect to the $L_\infty$-norm \cite{bonito2013adaptive}. Hence, if the permeability coefficients lie in compact subset of $L_\infty$, then the corresponding class of solution operators is compact. This can be applied as follows. Suppose that a set of forcing terms $f$ and their corresponding solutions $u$ are observed for an unknown permeability $k$. Using the observed data as in-context data, the network would be able to predict the solution for a new forcing term $f$. We demonstrate this application in Section \ref{subsec:darcy-experiments}.

\textbf{Finite unions of compact sets of operators.} Given any finite collection of sets of compact operators, their union is again compact and in-context multi-operator learning can be applied. For example, instead of considering only a single parameterized PDE as in the previous paragraph, multiple parameterized PDEs can be learned at the same time. We demonstrate this application in Section \ref{subsec:multie-differential-operators}.

\subsection{Universality for in-context operator learning}

We now formulate what it means for a neural network architecture to be universal for in-context operator learning. Roughly speaking, this means that the network can learn any compact set of operators to any desired accuracy provided enough in-context examples.

Recall that, although the grids $\textbf{x}$ and $\textbf{y}$ are fixed, we want to allow the in-context examples to be given for arbitrary functions $u_1,...,u_m\in V$ (even the number of functions $m$ should also be arbitrary). Of course, we need to make some assumption which guarantees that the $u_i$ sufficiently cover the domain $V$ so that the desired operator $G\in \mathcal{K}$ can be determined to the desired accuracy from the outputs $G(u_i)$. For technical reasons, we also need to choose the $u_i$ to be sufficiently uniformly distributed within $V$. This is captured in the following definition.
\begin{definition}[$(\delta,C)$-discretization]\label{delta-c}
Let $\delta > 0$ and $C > 1$. A $(\delta,C)$-discretization of a compact metric space $V$ is a set of points  $\textbf{u} = \{u_1, \cdots, u_m\}$ such that $$|\{u_i \in B(u,\delta/2)\}|<C|\{u_i \in B(v,\delta)\}|$$ for any $u, v \in V$.
\end{definition}
Intuitively, a $(\delta,C)$-discretization is a $\delta$-net of $V$ which is not too unbalanced, in the sense that the density of points is relatively uniform. Let us first note that this definition is not vacuous. 
\begin{lemma}\label{lem:discretization}
    Let $V$ be a compact set in a metric space, $\delta > 0$ and $C > 1$. Then there always exists a $(\delta,C)$-discretization of $V$.
\end{lemma}

We can now formulate the concept of universality for in-context multi-operator learning. 
\begin{definition}[Mathematical formulation of In-Context Operator Learning]\label{universality-definition}
    We define universality for in-context operator learning to mean that for any $\mathcal{K}\subset \mathcal{O}$, $\epsilon > 0$ and $C > 1$, there exist grids $\textbf{x} = (x_1,...,x_k)$ and $\textbf{y} = (y_1,...,y_n)$ in $K_1$ and $K_2$, respectively, a $\delta > 0$ and a network $K_\epsilon$ such~that
\begin{equation}
    |K_{\varepsilon}(u_1(\textbf{x}),G(u_1)(\textbf{y}),...,u_m(\textbf{x}),G(u_m)(\textbf{y}),u(\textbf{x}),y)- G(u)(y)| \,<\, \varepsilon\,,
\end{equation}
for any $G\in \mathcal{K}$, $u\in V$, $y\in K_2$, and $(\delta,C)$-discretization $\textbf{u} = (u_1,...,u_m)$ of $V$. In other words, any operator of $\mathcal{K}$ can be approximated to arbitrary accuracy provided that the in-context examples are sufficiently dense in $V$.
\end{definition}
Note that the network architecture should be able to process an arbitrary, i.e., not a priori specified, number of in-context examples. This can be achieved using a transformer architecture \cite{vaswani2017attention}, for example. In the next section, we describe a simple architecture based upon Deep Sets \cite{zaheer2017deep}, called DeepOSets \cite{chiu2024deeposets} for this purpose.

\section{The DeepOSets Architecture for In-Context Operator Learning}\label{model-section}

\begin{figure*}[tb]
    \includegraphics[width=\textwidth]{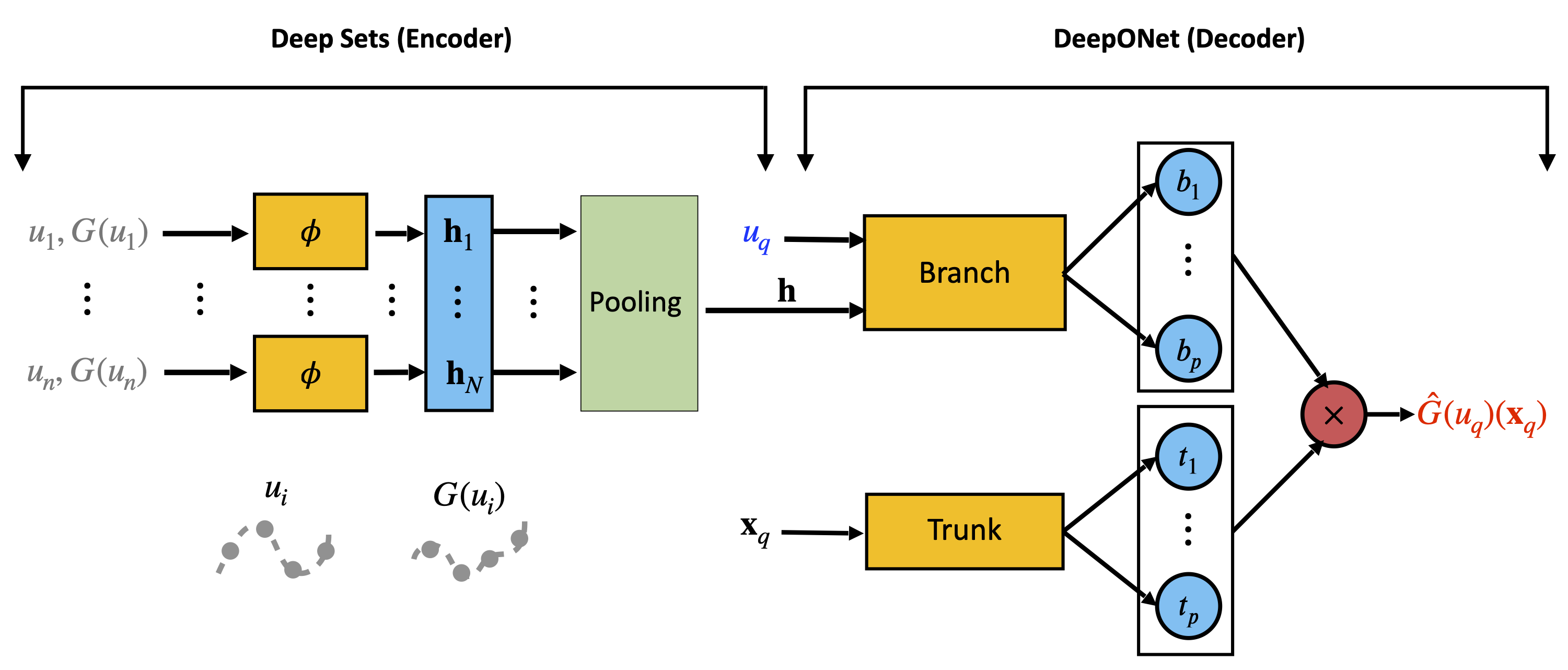}
    \caption{DeepOSets architecture for in-context multi-operator learning.}
    \label{fig:model}
\end{figure*}

The DeepOSets model is a hybrid neural network comprising of a DeepSets encoder and a DeepONet decoder, as seen in Figure~\ref{fig:model}.

%\subsection{DeepSet as encoder for in-context operator data}
The DeepSets module encodes the discretized parameter and solution function example pairs in the prompt. Each pair $(u_i, G(u_i))$ is first mapped to an individual fixed-length embedding $\bf{h}_i$ by a shared multi-layer perceptron (MLP) denoted $\Phi$. These intermediate representations are then combined by a pooling layer into the embedding vector $\bf{h}$. This introduces a permutation invariance inductive bias to the prompt examples, as long as the pooling layer is a permutation invariant operation, such as min, max, mean or set attention pooling. The permutation invariance is a key inductive bias, since the order of the in-context examples does not matter.

%The set of these embedding $\{\vec{h}_1, \dots, \vec{h}_n\}$ is aggregated using a pooling function to yield the final operator embedding $\vec{h}$. 
%\subsection{DeepONet as predictor for an operator}
The DeepONet takes the concatenation of the embedding $\bf{h}$ and the discretized query parameter function $u_q$ at its branch path and the query coordinates $x_q$ at its trunk path, and  outputs the predicted solution function $\hat{G}(u_q)$ evaluated at $x_q$. The output is therefore entirely mesh-free. We denote the branch network as $\rho$, and the trunk network as $\Psi$. With this notation, the output of the DeepOSets architecture using mean pooling, given input $x_q$, can be mathematically expressed as,
\begin{equation}
    \hat{G}(u_q)(x_q) \,:=\, \Psi(x_q)\cdot
    \rho\left(u_q(\textbf{x}),\frac{1}{m}\sum_{i=1}^m\Phi(u_i(\textbf{x}), G(u_i)(\textbf{y}))\right).
\end{equation}

\section{Universality of DeepOSets for In-Context Operator Learning}\label{sec:universality for in-context}
In this section, we prove that DeepOSets is a universal {\em uniform} approximator for any compact set of continuous operators in the sense of Definition \ref{universality-definition}, i.e., a single DeepOSets neural network exists that bounds the supremum of the error over the entire class to any desired fixed accuracy. This shows that a unique DeepOSets neural network could be learned that approximates in-context any continuous operator to the desired accuracy, given an appropriate number of examples in the prompt.

\begin{theorem}[Uniform universality for In-Context Operator Learning]\label{main-universality-theorem}
Given $\varepsilon >0$, $C > 1$ and a compact set $\mathcal{K}\subset \mathcal{O}$, there exist
integers $N,M \geq 1$, $\delta > 0$, fixed discretizations $\textbf{x} = \{x_1,...,x_k\}$ of $K_1$ and $\textbf{y} = \{y_1,...,y_n\}$ of $K_2$, and continuous functions $\Psi:\mathbb{R}^d\rightarrow \mathbb{R}^M$, $\rho:\mathbb{R}^{k+N}\rightarrow \mathbb{R}^M$, and $\Phi:\mathbb{R}^{k+n}\rightarrow\mathbb{R}^N$, such that 
\begin{equation}
    \sup_{G \in \mathcal{K}}\:\Bigg|\, G(u_q)(x_q)~-~
    \left.\Psi(x_q)\cdot\rho\left(u_q(\textbf{x}),\frac{1}{m}\sum_{i=1}^m\Phi(u_i(\textbf{x}), G(u_i)(\textbf{y}))\right)\right|<\varepsilon.
\end{equation}
for any $(\delta,C)$-discretization $\textbf{u} := (u_1, \cdots, u_m)$ of $V$, any $u_q\in V$, and any $x_q\in K_2$.
\end{theorem}
\begin{remark}
    Note that constructing a $(\delta,C)$-discretization may be difficult in practice. In addition, a $(\delta,C)$-discretization may need to be very large, which would require a large number of in-context examples. However, the universality result is an existence result, and in practice a much smaller number of in-context examples may be sufficient to obtain good accuracy.
\end{remark}
\begin{proof}
    For the full detailed proof, please see the Appendix.  %\ref{proof-of-universality}
    
    Here, we present an overview of the proof, in which the approximation error is decomposed into two inequalities. The compactness assumption enables this decomposition to hold uniformly over the class of continuous operators by reducing the problem to finite-dimensional approximations via discretizations on the grids $\mathbf{x}$ and $\mathbf{y}$.

    For the first inequality, using a partition of unity Lemma (in Appendix), we show that there exists a grid $\textbf{y}$ of $K_2$, and $\{\gamma_i\}_{i=1}^n \subset C(K_2)$, such that
    $$\|G(u_q) - \sum_{i=1}^nG(u_q)(y_i)\gamma_i\|_{C(K_2)} < \varepsilon/2,$$
    for every $G\in \mathcal{K}$ and $u_q\in V$.
    %\ref{proof-of-universality}
    
    For the second inequality, we show that there exists a grid $\textbf{x}$ of $K_1$ such that $G(u_q)(\textbf{y}) := (G(u_q)(y_1), \cdots, G(u_q)(y_n))$ can be approximated well by functions of the form,
    \begin{equation}\label{rho-eqn}
    \rho\left(u_q(\textbf{x}),\frac{1}{m}\sum_{i=1}^m\Phi(u_i(\textbf{x}), G(u_i)(\textbf{y}))\right),
    \end{equation}
    where $u_1,...,u_m$ is a $(\delta,C)$-discretization of $V$. 
    
    We construct $\Phi$ using a partition-of-unity argument to be localized/subordinate over a discretization of the input space, and define $\rho$ by forming normalized local averages of the corresponding operator output samples evaluated on the output grid and recombining them into a continuous finite-dimensional approximation.
    
    We then let $\Psi:\mathbb{R}^{d}\rightarrow\mathbb{R}^n$, (here $n=M)$, be defined as,
    $$\Psi(x) = \{\gamma_i(x)\}_{i=1}^n.$$
    Taking the dot product with \eqref{rho-eqn} and applying the triangle inequality together with the two bounds yields the desired approximation inequality.
    
    Finally, we verify that the constructions $\Phi$ and $\rho$ are continuous. Continuity of $\Phi$ is immediate from its definition, while continuity of $\rho$ follows from uniform lower bounds on the normalization factors ensured by the balanced $(\delta,C)$-discretization of $V$.
\end{proof}

The continuous functions guaranteed by Theorem~\ref{main-universality-theorem}
can be approximated using neural networks by invoking standard universal approximation theorems for MLPs \cite{cybenko1989approximation,hornik1991approximation,pinkus1999approximation}. Here it is important that we are averaging the outputs of $\Phi$ to ensure that the function $\rho$ only needs to be approximated on a compact set.

\section{Experimental Results}\label{sec:experiments}
\label{sec:experimental}
In this section, we demonstrate the in-context operator learning paradigm on the one-dimensional Burgers' equation, one and two dimensional Darcy flow, and a collection of different forward and inverse differential equation operators. We do not compare our method against single-operator learning methods, since our multi-operator DeepOSets has the fundamentally different goal of learning a compact class of operators with no knowledge of the underlying PDE parameters, which are never passed to the network. We also provide a controlled comparison with the ICON model \cite{yang2023icon,yang2024pde}. In contrast to ICON, we remark that our method is capable of directly handling two dimensional parametric PDEs, such as the two-dimensional Darcy flow in Section \ref{subsec:darcy-experiments2D}. The data generation process for all experiments is described in Appendix \ref{experimental-appendix}. 

\subsection{Burgers' equation}
First, we study Burgers' equations, 
\begin{align}\label{eq:burger}
    \frac{\partial u}{\partial t}+ u \frac{\partial u}{\partial x} &= \nu\frac{\partial^2u}{\partial x^2}, 
    \\
    u(x,0) &= f(x)
\end{align}
on $(x,t) \in [0,1]^2$, with $f(0)=f(1)=0$.

For a given $\nu$, let $G_{\nu}$ represent the solution operator for \ref{eq:burger}, which given an initial condition $u(x,0)$, returns the final time solution $u(x,1)$. The class of operators we seek to approximate can then be described as,
\begin{equation}\label{eq:burger-class}
    \{G_{\nu} : \nu \in [\alpha, \beta]\}
\end{equation}
for constants $\alpha$, $\beta$. For our experiments, we let $\nu \in [0.001, 0.1]$, and for data generation used log-uniform sampling to select $\nu$. 

Log-uniform sampling is important to mention, since for Burgers' equation, as $\nu$ tends towards 0, the solution will develop shocks, but for large $\nu$ the solution will be smooth, and so we need our generated training data to include adequate examples of both. For each sampled $\nu$, we generate 40 initial conditions $u_0$, and corresponding FDM solutions $u$.

During training, the network is given 39 in-context examples, and predicts the 40th one. At testing time we vary the number of in-context examples given from 3 to 39, and test 100 different operators, randomly selected from our test dataset, to calculate the relative $L_2$ error. 

\begin{figure}
    \begin{subfigure}[b]{0.33\textwidth}
        \includegraphics[width=\linewidth]{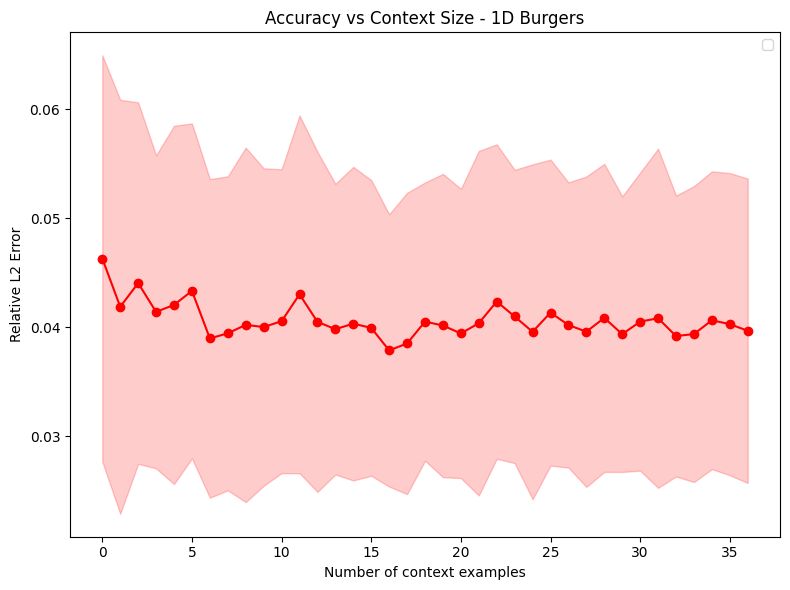}
        \label{fig:exp-frd}
    \end{subfigure}
    \begin{subfigure}[b]{0.33\textwidth}
        \includegraphics[width=\linewidth]{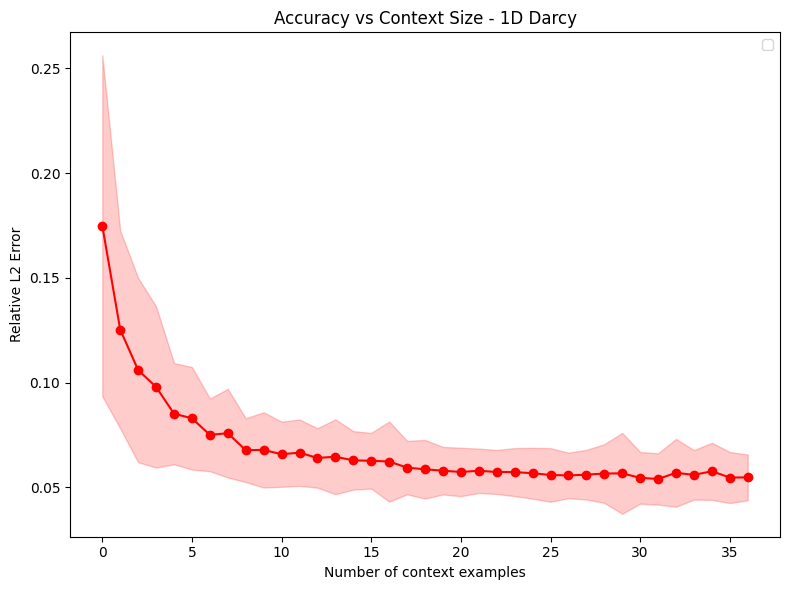}
        \label{fig:exp-inv}
    \end{subfigure}
    \begin{subfigure}[b]{0.33\textwidth}
        \includegraphics[width=\linewidth]{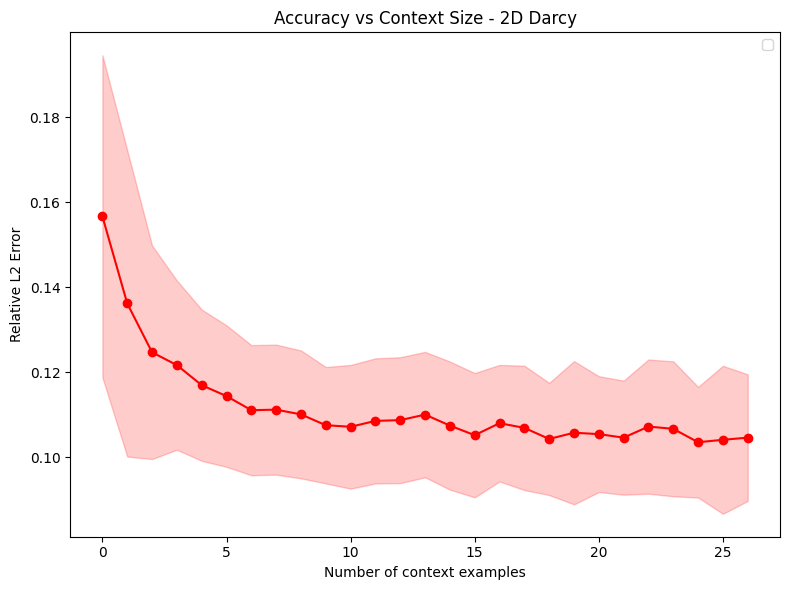}
        \label{fig:exp-inv2}
    \end{subfigure}
    \caption{\color{black}Relative $L_2$ error vs in-context examples for 1D Burgers', 1D Darcy Flow, and 2D Darcy Flow, with red shading representing one standard deviation, all experiments show a decrease in relative error as context increases, however it is much more pronounced in the Darcy examples, perhaps due to the more complicated operator paramatrization.}
    \label{fig:rel_vs_ex}
\end{figure}

As illustrated in \ref{fig:rel_vs_ex}, the relative error even with just 3 in-context examples is 0.0463, and this reduces to 0.0397 for 39 in-context examples. The overall small reduction with increasing in-context size, and relatively high variance, suggests that the DeepOSets architecture is able to identify the underlying operator with relatively few in-context examples. This intuitively makes sense for the class of Burgers' operators \ref{eq:burger-class}, since it is parametrized by a varying constant $\nu$ instead of a more complex function. Experimentally, our architecture is able to capture both smooth solutions, and developing shock solutions of Burgers', with only a few in-context examples as illustrated in \ref{fig:burger}.

\subsection{1D Darcy flow}
\label{subsec:darcy-experiments}

%\subsubsection{1D Problem}
We study the 1D Darcy flow, 
\begin{align}
-\frac{d}{dx}(k(x) \frac{d}{dx}u(x)) &= f(x), x \in (0,1) \\
u(0) &= u(1) = 0    
\end{align}
Here, given some $k$, our solution operator can be described as $G_k$. We generate multiple smooth $k's$, and for each one generate 40 functions $f$, and the corresponding FDM solution for $u$, which will then serve as our data. The training and testing were conducted in the same manner as Burgers', in terms of in context examples provided during training/testing, and how average $L_2$ error was calculated.
For 1D Darcy flow, we see a much more drastic decrease in average relative $L_2$ error on the test dataset, \ref{fig:rel_vs_ex}, with smaller variance, as the in-context examples increase, from 0.1749 to 0.0547. Since in this case, the class of operators is parametrized by a function $k(x)$, as opposed to a constant like in Burgers', the network greatly benefits from added in-context, as seen in \ref{fig:darcy1d}, and with enough context can perform well experimentally. Additional results can be found in Appendix~\ref{additional-experiments-appendix}.

\subsection{2D Darcy Flow}
\label{subsec:darcy-experiments2D}

We now study the 2D Darcy flow,
\begin{align}
    -\text{div}(k(x)\cdot \nabla u(x)) &= f(x), \ x \in (0,1)^2 := \Omega \\
    u(x) &= 0, x \ \in \partial\Omega
\end{align}
The class of operators can be described in the exact same manner as in 1D, and the data generation, training, and testing, all follow similar protocols as in 1D Darcy flow and Burgers' equation. The relative L2 vs.\ in-context examples can be seen in \ref{fig:rel_vs_ex}. Again, similarly to 1D Darcy, we see a noticeable reduction in relative $L_2$ error as the number of in-context examples increases, however this time it reduces from 0.1567 to 0.1046. Given the fact that the $k(x)$ functions can be much more complicated in the 2D case as compared to the 1D case, this is expected. Representative example solutions can be found in Appendix~\ref{additional-experiments-appendix}. However, the reduction in error does demonstrate the network is understanding how to disambiguate the operator. 

\subsection{Multiple differential equation operators}\label{subsec:multie-differential-operators}
A union of compact sets is also compact, and so it stands to reason that DeepOSets could feasibly learn multiple different operators for multiple PDEs with unknown parameters. In this section, we demonstrate that the DeepOSets architecture successfully learns the solution operator for various initial-value and boundary-value problems. Our evaluation spans 12 distinct tasks, comprising forward and inverse versions of each problem. The problems are parameterized differential equations, and our compact set in this case consists of the union of the operators obtained for various values of the parameters. See Table \ref{tbl:probs} for a summary of all the problems and the results. All of the implementation details and details on the training of the network can be found in Appendix~\ref{experimental-appendix}.

\begin{table*}[htp]
    \centering 
    \small
    \begin{tabular}{lccc}
    \textbf{Problem} & \textbf{Equations} & \textbf{Parameters} & \textbf{Test MSE}\\ 
    \hline
    Forward ODEConst & $\frac{du(t)}{dt} = a_1 c(t) + a_2$ & $a_1, a_2$ & 2.65e-03 $\pm$ 9.30e-03\\ 
    Inverse ODEConst &  &  & 1.15e-02 $\pm$ 2.71e-02\\ 
    Forward ODE1 & $\frac{du(t)}{dt} = a_1 u(t)c(t) + a_2$ & $a_1, a_2$& 1.65e-03 $\pm$ 8.39e-03\\ 
    Inverse ODE1 &  & & 1.44e-02 $\pm$ 3.86e-02\\ 
    Forward ODE2 & $\frac{du(t)}{dt} = a_1 u(t) + a_2 c(t) + a_3$ & $a_1, a_2, a_3$ & 1.19e-02 $\pm$ 6.99e-02\\ 
    Inverse ODE2 &  & & 1.40e-02 $\pm$ 9.15e-02\\ 
    Forward Poisson & $\frac{d^2 u(x)}{dx^2}=f(x)$  & None & 3.04e-03 $\pm$ 1.91e-02\\
    Inverse Poisson &  &  & 4.60e-03 $\pm$ 1.35e-02\\
    Forward Linear React-Diff & $-a \frac{d^2 u(x)}{dx^2} + k(x)u(x)=c$ & $u(0),u(1),a,c$& 2.11e-03 $\pm$ 4.25e-03\\
    Inverse Linear React-Diff & & & 1.45e-02 $\pm$ 6.01e-02\\
    Forward Nonlinear React-Diff & $-a\frac{d^2 u(x)}{dx^2}+ku(x)^3=c(x)$ &  $u(0),u(1),a,c$& 2.77e-03 $\pm$ 5.69e-03\\
    Inverse Nonlinear React-Diff & & & 1.19e-02 $\pm$ 3.51e-02\\
    \hline
    \end{tabular}
    \vspace{2ex}
    \caption{Summary of tasks. The reported Mean Squared Error (MSE) corresponds to the model trained on all 12 problems simultaneously. All models were tested on individual tasks.}
    \label{tbl:probs}
\end{table*}

\subsection{Extrapolation to Prompt Sizes not Seen During Training.}
\begin{figure}
    \begin{subfigure}[t]{0.5\textwidth}
        \caption{\hspace{1.5em}Forward problems}
        \includegraphics[width=\linewidth]{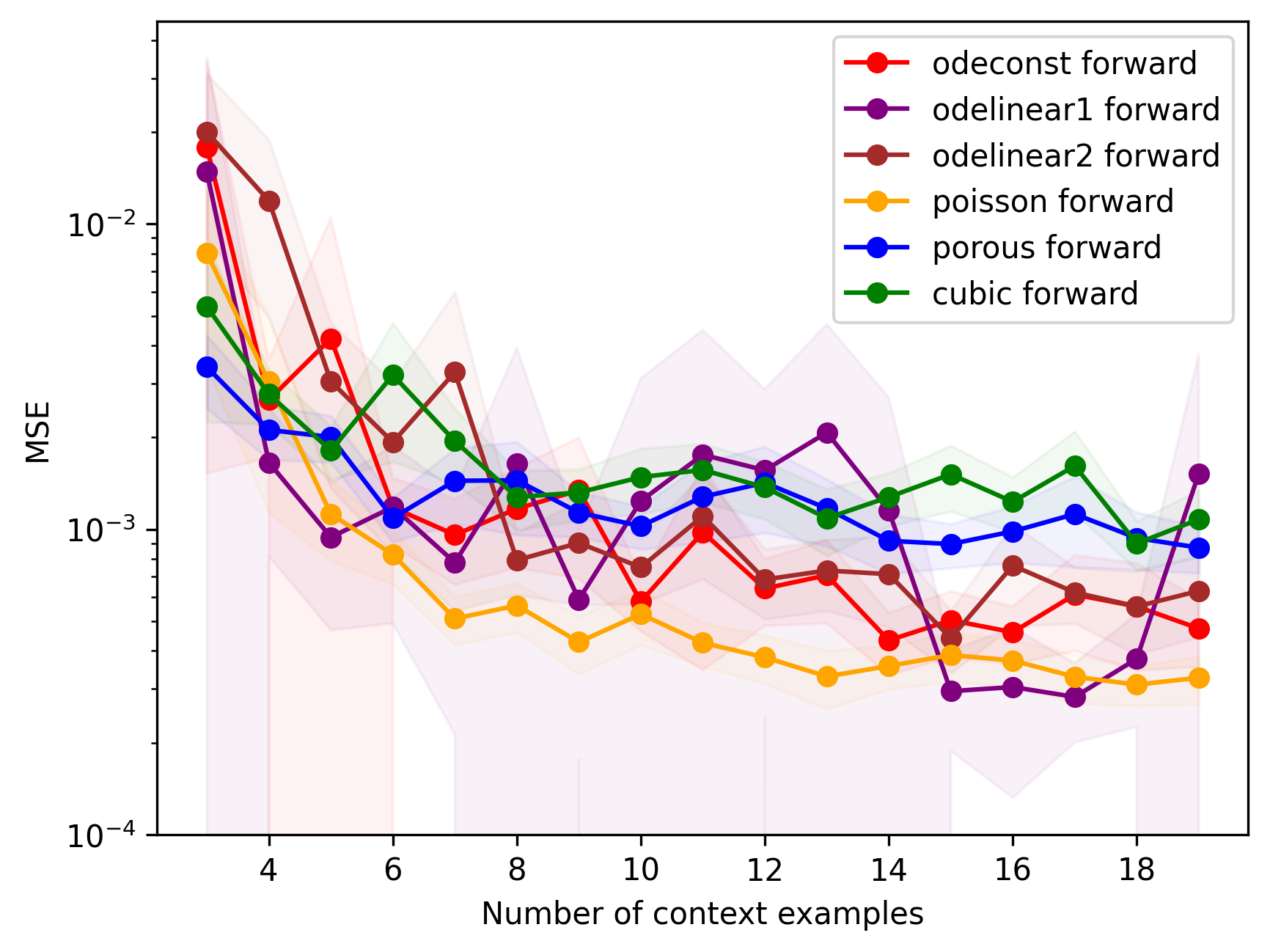}
        \label{fig:exp-frd2}
    \end{subfigure}
    \begin{subfigure}[t]{0.5\textwidth}
        \caption{\hspace{1.5em}Inverse problems}
        \includegraphics[width=\linewidth]{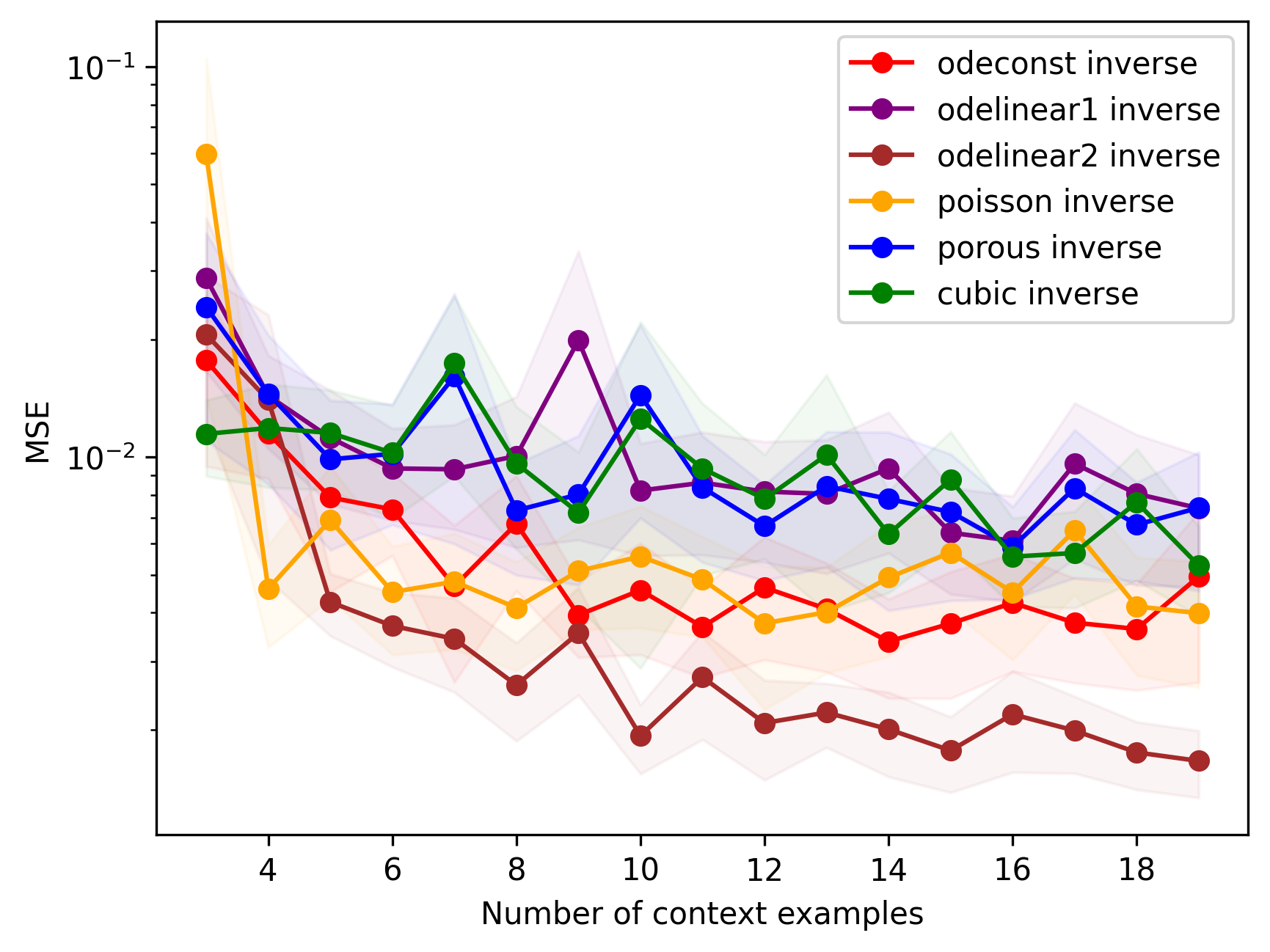}
        \label{fig:exp-inv3}
    \end{subfigure}
    \caption{\color{black}Zero-shot generalization to larger context sets. The model was trained with a fixed context size of $N=4$. During inference, we evaluate performance on context sizes ranging from 3 to 20. (a) Forward operators show a clear trend where providing more examples monotonically decreases the MSE, indicating the model successfully leverages additional data. (b) Inverse operators maintain stability across varying context sizes. Shaded regions denote one standard deviation.}
    \label{fig:exp}
\end{figure}

A distinct advantage of the DeepOSets framework is its permutation invariance and flexibility regarding the number of examples in the  prompt. To evaluate this extrapolation capability, we used different number of in-context examples ranging from $N=3$ to $N=20$ at inference time.
As illustrated in Figure~\ref{fig:exp}, the model successfully generalizes to context sizes significantly larger than the training distribution ($N_{\text{test}} > N_{\text{train}}$). Specifically, for the Poisson, ``porous'' (linear reaction diffusion) and ``cubic'' (nonlinear reaction diffusion) forward operators, we observe a consistent monotonic decrease in Mean Squared Error (MSE) as the number of context examples increases. This behavior confirms that the model is not overfitting to the specific training sequence length; rather, it effectively performs in-context learning, leveraging additional data points to refine the operator approximation. For the inverse problems, the model maintains performance stability across varying context sizes without divergence, demonstrating robust generalization.

%\section{Comparison with ICON and Ablation Studies}\label{app:icon-comparison}

% To position DeepOSets against an existing in-context multi-operator learning method, we carried out a controlled benchmark against ICON \cite{yang2023icon} on a 1D Darcy-flow problem, reported in this appendix. Under a short training budget (10k steps), DeepOSets attains lower error than ICON at every context size with 19\% of the parameters; with an extended budget (100k steps), ICON attains 26--41\% lower relative $L_2$ error, but at the cost of roughly 218$\times$ more training time, 12$\times$ higher inference latency, and 5.3$\times$ more parameters. This appendix also reports prompt ablations (no-context, mismatched, shuffled, and mispaired prompts) confirming that the model uses the prompt to select the operator, a nearest-neighbor baseline that performs 7--9$\times$ worse than DeepOSets, and a training-set-size ablation showing that the in-context performance of DeepOSets does not depend on a large training set.

\subsection{Comparison with ICON} %\label{subsec:icon-comparison}
To compare DeepOSets against an existing in-context multi-operator learning method, we carried out a controlled benchmark of DeepOSets and ICON \cite{yang2023icon} on a 1D Darcy flow (variable-coefficient Poisson) boundary-value problem, $-\frac{d}{dx}\big(k(x)\frac{du}{dx}\big) = f(x)$ on $(0,1)$ with $u(0)=u(1)=0$, where each operator corresponds to an independently sampled diffusivity field $k(x)$. The dataset consists of 5,000 operators with 40 $(f,u)$ pairs each, discretized on 50 interior grid points, with a 90\%/10\% operator-level train/test split. Full details of the dataset, architectures, and training protocol are given in Appendix \ref{app:icon-details}.

\begin{table}[htp]
    \centering
    \small
    \begin{tabular}{lrr}
    \toprule
    \textbf{Metric} & \textbf{DeepOSets} & \textbf{ICON} \\
    \midrule
    Trainable parameters & 894,465 & 4,744,193 \\
    Training time, 10k steps & 42.5 s & 2.15 h \\
    Training time, 100k steps & 144.6 s & 8.74 h \\
    Mean time per training step & 1.446 ms & 314.78 ms \\
    Training throughput & 691.7 steps/s & 3.18 steps/s \\
    Inference time per sample & 0.229 ms & 2.750 ms \\
    \bottomrule
    \end{tabular}
    \vspace{1ex}
    \caption{Computational cost of DeepOSets and ICON on the 1D Darcy flow benchmark. Inference time is the mean GPU wall-clock time per forward pass over 500 timed calls (100 per context size) for the models trained for 100k steps.}
    \label{tbl:icon-cost}
\end{table}

\begin{table}[htp]
    \centering
    \small
    \begin{tabular}{ccccc}
    \toprule
    & \multicolumn{2}{c}{\textbf{10k training steps}} & \multicolumn{2}{c}{\textbf{100k training steps}} \\
    \cmidrule(lr){2-3}\cmidrule(lr){4-5}
    $n_{\mathrm{ctx}}$ & DeepOSets & ICON & DeepOSets & ICON \\
    \midrule
    1 & \textbf{0.2831} & 0.7913 & 0.2096 & \textbf{0.1342} \\
    2 & \textbf{0.2085} & 0.8056 & 0.1180 & \textbf{0.0826} \\
    3 & \textbf{0.1881} & 0.6299 & 0.0976 & \textbf{0.0589} \\
    4 & \textbf{0.1740} & 0.6843 & 0.0821 & \textbf{0.0572} \\
    5 & \textbf{0.1459} & 0.7058 & 0.0819 & \textbf{0.0603} \\
    \bottomrule
    \end{tabular}
    \vspace{1ex}
    \caption{Mean relative $L_2$ test error of DeepOSets and ICON on the 1D Darcy flow benchmark as a function of the number of in-context examples $n_{\mathrm{ctx}}$, averaged over 100 held-out test samples per condition, under equal-step training schedules of 10k and 100k steps.}
    \label{tbl:icon-error}
\end{table}

The results are summarized in Tables~\ref{tbl:icon-cost} and \ref{tbl:icon-error}. Under a small training budget of 10,000 steps, DeepOSets outperforms ICON at every context size while using only 19\% of ICON's parameters; ICON requires substantially more training steps to converge, given its much larger sequence length and parameter count. When the training schedule is extended to 100,000 steps, ICON attains a lower mean relative $L_2$ error than DeepOSets at every tested context size, with a relative reduction in error ranging from approximately 26\% to 41\%. This accuracy comes at a cost: under the equal-step schedule, ICON required approximately 218 times more wall-clock training time than DeepOSets (8.74 hours versus 144.6 seconds), roughly 12 times higher per-sample inference latency, and about 5.3 times more parameters. The training-time gap is consistent with the cost of self-attention over a token sequence of length 850 at each step, in contrast to the DeepOSets encoder, whose cost is linear in the number of context pairs and which processes a short fixed-size input. In summary, the comparison reveals a clear trade-off between predictive accuracy and computational cost: ICON achieves better accuracy with extended training, while DeepOSets is substantially more computationally efficient and attains comparable accuracy at a small fraction of the cost.

\section{Discussion and Limitations}

A limitation of our work is that the universality result we have proved is not quantitative. In particular, we do not know how large the network must be, nor how many in-context examples are required, to learn a given compact class of operators. For operators arising from PDEs, this as an important direction for future research.
Our experimental results demonstrate the viability of DeepOSets on one and two-dimensional cases with both finite and infinite-dimensional unknown parameters. We also show that DeepOSets is competitive with ICON, a much more complex transformer-based architecture. While ICON is more accurate at 100K training steps, the difference with respect to DeepOSets is small at 5 in-context examples, while DeepOSets is much faster to train.

\section{Conclusion}

Multi-operator learning involves training a neural network to 
learn a class of operators simultaneously. In-context multi-operator learning follows an amortization approach, where a neural network is trained on a large number of examples sampled from the universe of operators and, at inference time, a prediction is produced quickly by disambiguating the correct operator using a small number of examples provided in a prompt, without any neural network weight updates. We have studied\linebreak in-context multi-operator learning through the DeepOSets architecture, which we prove to be a universal approximator over a compact class of operators and demonstrate to be viable in practice through a comprehensive set of experiments with operators arising from PDEs. Pushing the boundaries of what is possible using this approach is an important, and promising, area of research.

\bibliography{refs}
\bibliographystyle{plain}

\appendix

\section{Proofs of Theorem \ref{main-universality-theorem} and Lemma \ref{lem:discretization}}
Let us begin by proving Lemma \ref{lem:discretization} showing that $(\delta,C)$-discretizations always exist for any compact metric space $V$.
\begin{proof}[Proof of Lemma \ref{lem:discretization}]\label{proof-of-discretization}
Let $V \subset X$ be compact, and let $\delta>0$ be given. Choose a maximal $\delta$-separated subset of $V$, i.e., a subset $\mathcal{X}_\delta$ such that $d(u_i,u_j) \geq \delta$ for all $i\neq j$, and which is maximal with respect to this property. It is clear that the balls of radius $\delta$ around the points in $\mathcal{X}_\delta$ give a covering of $V$ (since any point not covered by such a ball could be added to $\mathcal{X}_\delta$, contradicting the maximality), and so by compactness $\mathcal{X}_\delta$ must be a finite set. We claim that $\mathcal{X}_\delta$ is a $(\delta,C)$-covering for any $C > 1$.

First, note that any ball of radius $\delta/2$ contains at most one point of 
$\mathcal{X}_\delta$.  Indeed, fix $z \in V$ and suppose 
$u_i, u_j \in \mathcal{X}_\delta$ both lie in $B(z,\delta/2)$ with $i \neq j$.  
Then, by the triangle inequality,
\[
  d(u_i,u_j) 
  \;\le\; d(u_i,z) + d(z,u_j) 
  \;<\; \frac{\delta}{2} + \frac{\delta}{2} 
  \;=\; \delta,
\]
which contradicts the $\delta$--separation of $\mathcal{X}_\delta$.  
Hence, for every $z\in V$,
\[
  \bigl|\{u_i \in \mathcal{X}_\delta : u_i \in B(z,\delta/2)\}\bigr| \le 1.
\]

Next, since balls of radius $\delta$ around the points in $\mathcal{X}_\delta$ cover $V$,
for every $v\in V$ there exists some $u_j \in \mathcal{X}_\delta$ with 
$d(v,u_j) < \delta$, i.e.\ $u_j \in B(v,\delta)$.  Thus, for all $v\in V$,
\[
  \bigl|\{u_i \in \mathcal{X}_\delta : u_i \in B(v,\delta)\}\bigr| \ge 1.
\]
Combining these two observations, we obtain, for all $u,v\in V$,
\[
  \bigl|\{u_i \in \mathcal{X}_\delta : u_i \in B(u,\delta/2)\}\bigr|
  \;<\;
  C\,\bigl|\{u_i \in \mathcal{X}_\delta : u_i \in B(v,\delta)\}\bigr|,
\]
for any $C>1$.  
Hence $\mathcal{X}_\delta$ is a
$(\delta,C)$--discretization of $V$.
\end{proof}

Next, we state a partition of unity approximation Lemma for a compact subset of $C(K)$, which we will use in the proof of Theorem \ref{main-universality-theorem}.

\begin{lemma}\label{partition-of-unity-lemma}
Let $K$ be a compact metric space and $V \subset C(K)$ be compact. Given $\varepsilon > 0$, there exist $\mathbf{y} := \{y_1, \cdots,y_n\} \in K$, and $\varphi_1, \cdots, \varphi_n \in C(K)$ with $\|\varphi_i\| \leq 1$, such that
$$\|f - \sum_{i=1}^nf(y_i)\varphi_i\|_{C(K)} < \varepsilon$$
for all $f \in V$. In particular, for any $f,g\in V$ we have
\begin{equation}
    \|f - g\|_{C(K)} \leq \|f - g\|_{\ell^\infty(\mathbf{y})} + 2\varepsilon.
\end{equation}
\end{lemma}
\begin{proof} (of Lemma \ref{partition-of-unity-lemma})
Since $V$ is compact, by the Arzela-Ascoli theorem it is also equicontinuous, and so there exists $\delta > 0$ such that if $|x-y| < \delta$, then $|f(x) - f(y)| < \varepsilon$ for all $f \in V$. Since $K$ is also compact, we can pick $y_1, \cdots, y_n \in K$ such that $\{B(y_i, \delta)\}_{i=1}^n$ form a covering of $K$. Finally, let $\{\varphi_i\}_{i=1}^n$ be a partition of unity subordinate to the balls $\{B(y_i, \delta)\}_{i=1}^n$. 
\\\\
Given $y \in K$, for any $f\in V$, we have that,
\begin{align*}
    |f(y) - \sum_{i=1}^n f(y_i)\varphi_i(y)| &= |\sum_{i=1}^n (f(y) -  f(y_i))\varphi_i(y)| \\
    &<|\sum_{i, |y-y_i|<\delta} (f(y) -  f(y_i))\cdot\varphi_i(y)| \\
    &<|\sum_{i, |y-y_i|<\delta} \varepsilon \cdot \varphi_i(y))| \leq \varepsilon.
\end{align*}
Using this, if $f,g\in V$ we have
\begin{equation}
\begin{split}
    \|f - g\|_{C(K)} &\leq \|f - \sum_{i=1}^nf(y_i)\varphi_i\|_{C(K)} + \|f-g\|_{\ell^\infty(\mathbf{y})}\max_i \|\varphi_i\|_{C(K)} + \|g - \sum_{i=1}^ng(y_i)\varphi_i\|_{C(K)}\\
    &\leq \|f-g\|_{\ell^\infty(\mathbf{y})} + 2\varepsilon.
\end{split}
\end{equation}
\end{proof}

We can now return to the proof of Thm \ref{main-universality-theorem}.
\begin{proof} (of Theorem \ref{main-universality-theorem}) \label{proof-of-universality}
Fix $\varepsilon > 0$. Since $\mathcal{K}$ is compact and $V$ is compact, it follows that the set
\begin{equation}
    \{G(u):~G\in \mathcal{K},~u\in V\}\subset C(K_2)
\end{equation}
is compact.
Since $K_2$ is also compact, we can apply Lemma \ref{partition-of-unity-lemma} to get that there exists a grid $\textbf{y}$ of $K_2$, and $\{\gamma_i\}_{i=1}^n \subset C(K_2)$, such that
$$\|G(u_q) - \sum_{i=1}^nG(u_q)(y_i)\gamma_i\|_{C(K_2)} < \varepsilon/2,$$
for every $G\in \mathcal{K}$ and $u_q\in V$.
So let $\Psi:\mathbb{R}^{d}\rightarrow\mathbb{R}^n$ be defined as,
$$\Psi(y) = \{\gamma_i(y)\}_{i=1}^n.$$
Using the notation $G(u_q)(\textbf{y}) := (G(u_q)(y_1), \cdots, G(u_q)(y_n))$, we have that
\begin{equation}\label{y-discretization-equation}
    \|G(u_q) - G(u_q)(\textbf{y}) \cdot \Psi\|_{C(K_2)} < \varepsilon/2.
\end{equation}
for every $G\in \mathcal{K}$.

We now construct a grid $\textbf{x} = (x_1,...,x_k)$ in $K_1$ and continuous functions $\Phi$ and $\rho$, such that $G(u_q)(\textbf{y})$ is approximated to accuracy $\varepsilon/2$ by
\begin{equation}
    \rho\left(u_q(\textbf{x}),\frac{1}{m}\sum_{i=1}^m\Phi(u_i(\textbf{x}), G(u_i)(\textbf{y}))\right),
\end{equation}
for any $(\delta,C)$-discretization $u_1,...,u_m$ of $V$. 
\\\\
Since $\mathcal{K}$ is compact, it is equi-continuous, so there exists a $\delta$ such that if $\|u - v\|_{C(K_1)} < 37\delta/6$, then $\|G(u) - G(v)\|_{C(K_2)} < \varepsilon/(2\cdot \|\Psi\|_{C(K_2)})$ for every $G\in \mathcal{K}$. By Lemma \ref{partition-of-unity-lemma}, there exists $\textbf{x} = \{x_1, \cdots, x_k\} \in K_1$ such that
\begin{equation}\label{near-discretization-equation}
    \|u - v\|_{C(K_1)} \leq \|u - v\|_{\ell^\infty(\mathbf{x})} + \frac{\delta}{6}.
\end{equation}
We let $\textbf{x}$ be our grid in $K_1$ and $\delta$ be the value given in the Theorem.

Let $u_1,...,u_m\in V$ be a $(\delta,C)$-discretization of $V$, and consider the restrictions $u_i(\mathbf{x})$ of $u_i$ to the grid $\mathbf{x}$, which lie in the compact set $V(\mathbf{x})\subset \mathbb{R}^k$. Let $u\in K$ and observe that by \eqref{near-discretization-equation} we have
\begin{equation}
    \|u - u_i\|_{C(K_1)} - \frac{\delta}{6}\leq \|u - u_i\|_{\ell^\infty(\mathbf{x})} \leq \|u - u_i\|_{C(K_1)}
\end{equation}
This implies that
\begin{equation}
    \{i:u_i(\mathbf{x})\in B(u(\mathbf{x}),\delta/3)\} \subset \{i:u_i\in B(u,\delta/2)\} 
\end{equation}
and that
\begin{equation}
    \{i:u_i\in B(u,\delta)\} \subset \{i:u_i(\mathbf{x})\in B(u(\mathbf{x}),\delta)\}.
\end{equation}
Hence, since $u_1,..,u_m$ is a $(\delta,C)$-discretization, when restricted to the grid $\mathbf{x}$ we still have
\begin{equation}\label{modified-delta-C-discretization-equation}
    |\{u_i(\mathbf{x})\in B(u(\mathbf{x}),\delta/3)\}|\leq C|\{u_i(\mathbf{x})\in B(v(\mathbf{x}),\delta)\}|
\end{equation}
for any $u,v\in V$.
\\\\
Next, we choose a maximal $3\delta$-separated subset $\widetilde{u}_1(\mathbf{x}),...,\widetilde{u}_l(\mathbf{x})$ of the set $V(\mathbf{x})\subset \mathbb{R}^k$, and consider the open cover of $V(\mathbf{x})$ given by
\begin{equation}
    U_j := B(\widetilde{u}_j(\mathbf{x}),3\delta) \backslash\bigcup_{i=1}^j \overline{B(\widetilde{u}_i(\mathbf{x}),\delta)}.
\end{equation}
These sets are open since we are removing a finite union of closed sets from the open ball $B(\widetilde{u}_j(\mathbf{x}),3\delta)$. By maximality, the balls $B(\widetilde{u}_j(\mathbf{x}),3\delta)$ cover $V(\textbf{x})$, and so each point $v(\textbf{x})\in V(\textbf{x})$ is contained in $B(\widetilde{u}_j(\mathbf{x}),3\delta)$ for some $j$. If
\begin{equation}
    v(\textbf{x})\notin \bigcup_{i=1}^l \overline{B(\widetilde{u}_i(\mathbf{x}),\delta)},
\end{equation}
then $v(\textbf{x})\in U_j$. Otherwise, $v(\textbf{x})\in \overline{B(\widetilde{u}_i(\mathbf{x}),\delta)}$ for some $i$, and hence for exactly one $i$ since the balls $\overline{B(\widetilde{u}_i(\mathbf{x}),\delta)}$ are disjoint (because the $\widetilde{u}_i(\mathbf{x})$ are $3\delta$-separated). In this case, we have $v(\textbf{x})\in U_i$. Hence, the sets $U_j$ cover $V(\textbf{x})$.

Let $\varphi_j:\mathbb{R}^k\rightarrow \mathbb{R}_{\geq 0}$ be a partition of unity subordinate to the covering $U_j$. By construction, the balls of radius $\delta$ about $\widetilde{u}_j(\mathbf{x})$ are contained in $U_j$ and disjoint from $U_i$ for $i\neq j$, and so we have
\begin{equation}\label{partition=of-unity-lower-bound}
    \varphi_j(v(\mathbf{x})) = 1~~\text{whenever $\|v - \widetilde{u}_j\|_{\ell^\infty(\mathbf{x})} \leq \delta$.}
\end{equation}
In addition, by construction we have
\begin{equation}
    \varphi_j(v(\mathbf{x})) =0 ~~\text{whenever $\|v - \widetilde{u}_j\|_{\ell^\infty(\mathbf{x})} \geq 3\delta$.}
\end{equation}
Let $\Phi:\mathbb{R}^{k+n} \rightarrow \mathbb{R}^{l(n+1)}$ (so that $N = l(n+1)$) be defined by
$$\Phi(\mathbf{u}, \mathbf{v}) = \{\varphi_j(\mathbf{u})\cdot \mathbf{v}, \varphi_j(\mathbf{u})\}_{j=1}^{l}.$$
where $\mathbf{u} \in \mathbb{R}^k$ and $\mathbf{v} \in \mathbb{R}^n$. We then have for any $G\in \mathcal{K}$ and $u_1,...,u_m\in V$ that
\begin{equation}\label{Phi-output-equation}
    \frac{1}{m}\sum_{i=1}^m \Phi(u_i(\textbf{x}), G(u_i)(\textbf{y})) = \left\{  \frac{1}{m}\sum_{i=1}^m\varphi_j(u_i(\textbf{x}))\cdot G(u_i)(\textbf{y}), \frac{1}{m}\sum_{i=1}^m\varphi_j(u_i(\textbf{x}))\right\}_{j=1}^{l}.
\end{equation}
We claim that if $u_1,...,u_m$ is a $(\delta,C)$-discretization, then the sum
\begin{equation}
    \frac{1}{m}\sum_{i=1}^m\varphi_j(u_i(\textbf{x}))
\end{equation}
is bounded uniformly from below.

To prove this, let $v_1,...,v_t\in V$ be chosen so that $B(v_1(\textbf{x}),\delta/3),...,B(v_t(\textbf{x}),\delta/3)$ is a covering of $V(\textbf{x})$. It follows that one of these balls must contain at least $m/t$ elements $u_i(\textbf{x})$ (note that the size $m$ of the $(\delta,C)$-discretization is not controlled). Using \eqref{modified-delta-C-discretization-equation} this means each ball $B(u_i(\textbf{x}),\delta)$ must contain at least $m/(Ct)$ elements of the $(\delta,C)$-discretization. Finally, using \eqref{partition=of-unity-lower-bound} this gives
 \begin{equation}\label{normalizing-sum-lower-bound}
     \frac{1}{m}\sum_{i=1}^m\varphi_j(u_i(\textbf{x})) \geq \frac{1}{m}\frac{m}{Ct} = \frac{1}{Ct} > 0.
 \end{equation}

Now, define $\xi:\mathbb{R}^{l(n+1)} \rightarrow \mathbb{R}^{ln}$ to be
$$\xi(z_1, \cdots, z_{l(n+1)}) = \left\{ \frac{(z_j, z_{j+1}, \cdots, z_{j+n-1})}{z_{j+n}} \right\}_{j \in [1, (n+1)+1, 2(n+1)+1, \cdots, (l-1)(n+1)+1]},$$
when $|z_{j+n}| \geq 1/(Ct)$ for each $j$, and extend it continuously to the rest of $\mathbb{R}^{l(n+1)}$ using the Tietze extension theorem.
Then, given any input from \eqref{Phi-output-equation} where $u_1,...,u_m$ is a $(\delta,C)$-discretization, using \eqref{normalizing-sum-lower-bound} we will have
$$\xi\left(\frac{1}{m}\sum_{i=1}^m \Phi(u_i(\textbf{x}), G(u_i)(\textbf{y}))\right) = \left\{ \frac{\frac{1}{m}\sum_{i=1}^m\varphi_j(u_i(\textbf{x}))\cdot G(u_i)(\textbf{y})}{\frac{1}{m}\sum_{i=1}^m\varphi_j(u_i(\textbf{x}))} \right\}_{j=1}^{l} := \{\overline{G(\widetilde{u}_j)(\textbf{y})}\}_{j=1}^{l}.$$
Here the $\overline{G(\widetilde{u}_j)(\textbf{y})}$ notation is used just to emphasize that this is just the weighted average of the $G(u_i)(\textbf{y})$ samples that are in the ball of radius $\delta$ centered at $\widetilde{u}_j$. We then define, $\rho:\mathbb{R}^{k + l(n + 1)} \rightarrow \mathbb{R}^n$, as
\begin{equation}
    \rho(x,y) = \sum_{j=1}^l \textbf{(}\xi(y)\textbf{)}_j\cdot \varphi_j(x),
\end{equation}
so that we have
\begin{equation}\label{eqn:average-over-l-to-m}
    \rho\left(u_q(\textbf{x}),\frac{1}{m}\sum_{i=1}^m\Phi(u_i(\textbf{x}), G(u_i)(\textbf{y}))\right) = \sum_{j=1}^l \overline{G(\widetilde{u}_j)(\textbf{y})} \cdot\varphi_j(u_q(\textbf{x})).
\end{equation}
Now, fix any $G\in \mathcal{K}$. We will show that the above expression approximates $G(u_q)(\textbf{y})$. We calculate
\begin{align*}
    \left\|G(u_q)(\textbf{y}) -  \sum_{j=1}^l \overline{G(\widetilde{u}_j)(\textbf{y})} \cdot\varphi_j(u_q(\textbf{x}))\right\|_{\ell^\infty(\textbf{y})} &= \left\|\sum_{j=1}^l G(u_q)(\textbf{y})\cdot\varphi_j(u_q(\textbf{x})) -  \sum_{j=1}^l \overline{G(\widetilde{u}_j)(\textbf{y})} \cdot\varphi_j(u_q(\textbf{x}))\right\|_{\ell^\infty(\textbf{y})} \\
    & = \left\|\sum_{j, \|u_q-\widetilde{u_j}\|_{\ell^\infty(\textbf{x})}<3\delta} (G(u_q)(\textbf{y}) - \overline{G(\widetilde{u}_j)(\textbf{y}))}\cdot\varphi_j(u_q(\textbf{x}))\right\|_{\ell^\infty(\textbf{y})}
\end{align*}
Then, note
\begin{align*}
    \left\|G(u_q)(\textbf{y}) - \overline{G(\widetilde{u}_j)(\textbf{y})}\right\|_{\ell^\infty(\textbf{y})} &= \left\|G(u_q)(\textbf{y}) - \frac{\sum_{i=1}^m\varphi_j(u_i(\textbf{x}))\cdot G(u_i)(\textbf{y})}{\sum_{i=1}^m\varphi_j(u_i(\textbf{x}))} \right\|_{\ell^\infty(\textbf{y})} \\
    &= \left\| G(u_q)(\textbf{y}) - \sum_{i=1}^m \frac{\varphi_j(u_i(\textbf{x}))}{\sum_{i=1}^m \varphi_j(u_i(\textbf{x}))} \cdot G(u_i)(\textbf{y}) \right\|_{\ell^\infty(\textbf{y})} \\
    &= \left\| \sum_{i=1}^m \frac{\varphi_j(u_i(\textbf{x}))}{\sum_{i=1}^m \varphi_j(u_i(\textbf{x}))} \cdot\left( G(u_q)(\textbf{y}) - G(u_i)(\textbf{y})\right) \right\|_{\ell^\infty(\textbf{y})} \\
    &\leq \varepsilon/(2\cdot \|\Psi\|_{C(K_2)})\left| \sum_{i=1}^m \frac{\varphi_j(u_i(\textbf{x}))}{\sum_{i=1}^m \varphi_j(u_i(\textbf{x}))} \right| = \varepsilon/(2\cdot \|\Psi\|_{C(K_2)})
\end{align*}
Where we use that $\|u_q-\widetilde{u}_j\|_{\ell^\infty(\textbf{x})} < 3\delta$, $\|\widetilde{u}_j - u_i\|_{\ell^\infty(\textbf{x})} < 3\delta$, and \eqref{near-discretization-equation} imply that 
\begin{equation}
    \|u_q-u_i\|_{C(K_1)}\leq \|u_q-\widetilde{u}_j\|_{\ell^\infty(\textbf{x})} + \|\widetilde{u}_j-u_i\|_{\ell^\infty(\textbf{x})} + \delta/6 < 37\delta/6,
\end{equation}
so that by the choice of $\delta$, 
$\|G(u_q) - G(u_i)\|_{C(K_2)} <\varepsilon/(2\cdot \|\Psi\|_{C(K_2)})$.
\\\\
So we have 
$$\left\|G(u_q)(\textbf{y}) -  \sum_{j=1}^l \overline{G(\widetilde{u}_j)(\textbf{y})} \cdot \varphi_j(u_q(\textbf{x}))\right\|_{\ell^\infty(\textbf{y})} \leq \varepsilon/(2\cdot \|\Psi\|_{C(K_2)}),$$
which combining with \eqref{y-discretization-equation} and using (\ref{eqn:average-over-l-to-m}) gives us,
\begin{align*}
    &\|G(u_q) - \rho\left(u_q(\textbf{x}),\frac{1}{m}\sum_{i=1}^m\Phi(u_i(\textbf{x}), G(u_i)(\textbf{y}))\right) \cdot \Psi\|_{C(K_2)}  \\
    &\leq\|G(u_q) - G(u_q)(\textbf{y}) \cdot \Psi\|_{C(K_2)} + \|G(u_q)(\textbf{y}) \cdot \Psi - \rho\left(u_q(\textbf{x}),\frac{1}{m}\sum_{i=1}^m\Phi(u_i(\textbf{x}), G(u_i)(\textbf{y}))\right) \cdot \Psi\|_{C(K_2)}  \\
    &<\varepsilon/2 + \varepsilon/2 = \varepsilon.
\end{align*}
Since $\|.\|_{C(K_2)}:=\sup_{x \in K_2}\|.(x)\|$, this result then holds for any $x_q \in K_2$, as desired. This completes the proof.

% However, we still need to show that $\rho$ and $\Phi$ can be chosen continuous. It is clear from the definition that $\Phi$ is continuous. To show that $\rho$ is also continuous, we need to show that the denominator
% \begin{equation}
%     \frac{1}{m}\sum_{i=1}^m\varphi_j(u_i(\textbf{x}))
% \end{equation}
%  in the definition of $\xi$ is bounded away from zero uniformly for any $(\delta,C)$-discretization $u_1,...,u_m$ of $V$.
%  \\\\
%  To prove this, we use \eqref{modified-delta-C-discretization-equation} and \eqref{partition=of-unity-lower-bound}. Let 
\end{proof}

\section{Experimental Details}\label{experimental-appendix}

%or Burgers', and 1D/2D Darcy, to generate our training/testing data, we sample operators from a parametric family, and then solve the corresponding PDE numerically for various queries. We then split this into train/test sets.

\subsection{Burgers' equation}

We first sample 200,000 different $\nu$'s using log uniform sampling on the interval $[\log(0.001), \log(0.1)]$.
For each $\nu$, we generate 40 different initial conditions $u(x,0)$'s. To generate these, we draw from a Gaussian process with the radial basis function $K(x,x') = \sigma^2\exp(\frac{-(x-x')^2}{2\ell^2})$. We pick $\sigma = 0.8, \ell = 0.2$ to ensure our initial conditions have sufficient variability. We then apply a window function of $\sin^2(\pi x)$ to ensure the 0 boundary conditions are satisfied.

For a $\nu, u(x,0)$ pair, to solve the resulting PDE and get our reference solutions $u(x,1)$ we use an implicit explicit (IMEX) finite difference method (FDM), where we use forward difference in time, handle the nonlinear advection term $u\frac{\partial u}{\partial x}$ explicitly, and the diffusion term $\nu \frac{\partial^2 u}{\partial x^2}$ implicitly. We use 100 points to discretize the space $x \in [0,1]$, and 1000 points to discretize the time $t \in [0,1]$, to ensure shocks can be adequately captured.

Once the data is generated, we use a 90/10 training/testing split. During training, we use mini-batches of 5 operators, each with 39 in context examples (where the network predicts the 40th query). The architecture is trained for 100k iterations, using an Adam optimizer with a learning rate 1e-3. Training took approximately 12 minutes, using an Nvidia A100GPU via Google Colab, to achieve a loss of 3.5e-5.

The DeepOSets architecture has the following hyperparameters. For the DeepSets encoder, we use an MLP $\phi$ with 4 layers of 800 neurons each. The pooled latent space has dimension 800 with mean pooling. The branch and trunk networks of the DeepONet both have 5 layers of 800 neurons, and both embed into a dimension 800 space. All networks use ReLU activation functions. 

Once trained, we evaluate the model by calculating the relative L2 error for 100 randomly sampled operators form the test dataset, and repeat this for various in context sizes to see how the error changes. Comments on performance are found in $\ref{sec:experimental}$ 

\subsection{1D Darcy Flow}

%\subsubsection{1D Problem}
We first sample 200,000 different $k(x)$ coefficient functions to create our operators. For each $k(x)$, we generate 40 different forcing conditions (the RHS of the PDE) $f(x)$'s. 
To generate both the $k$'s and $f$'s, we draw from a Gaussian process with the radial basis function $K(x,x') = \sigma^2\exp\left(\frac{-(x-x')^2}{2\ell^2}\right)$. For $k$, we pick $\sigma = 0.5, \ell = 0.4$ to ensure our initial conditions have sufficient variability without extremely drastic functions. With similar reasoning, for $f$, we pick $\sigma = 2, \ell = 0.4$. These choices also guarantee a range of solution $u$ outputs of varying size and scale, which will help show our architecture can capture varying solution sizes.

For a $k, f$ pair, to solve the resulting PDE and get our reference solutions $u(x)$ we use a second order FDM scheme, where the diffusion coefficient is evaluated at cell interfaces using arithmetic averaging $(k_{i+1/2} = \tfrac{1}{2}(k_i + k_{i+1}))$. We use 50 points to discretize the space $[0,1]$.

Once the data is generated, we proceed similarly to the Burgers' case, using the same 90/10 split, training steps, and learning rate. We additionally use the same architeture hyperparameters as in Burgers', to show our architecture can learn various classes of operators without excessive tuning. Training again took 12 minutes, using an Nvidia A100GPU via Google Colab, to achieve a loss of 3.6e-5. 

Once trained, we evaluate the model by calculating the relative L2 error for 100 randomly sampled operators form the test dataset, and repeat this for various in context sizes to see how the error changes. Comments on performance are found in $\ref{sec:experimental}$

\subsection{2D Darcy Flow}
This experiment was run in a similar fashion to the 1D Darcy flow, however we used 50,000 operators, 40 examples for each operator, with a 90/10 split. Additionally for 2D problems, random Fourier feature approximations are used to sample smooth random fields to generate $k(x)$ and $f(x)$, avoiding the memory cost of the full GP covariance matrix. 

For a $k, f$ pair, to solve the PDE and get our reference solutions $u(x)$ we use an finite element method (FEM), using a square mesh with bilinear quadrilateral elements. We use a $32 \times 32$ discretization of the space.
We again train for 100k steps using Adam with a learning rate of 1e-3, to achieve a loss of 3.7e-5. Training took only 16 minutes, using an Nvidia A100GPU via Google Colab, despite the 2D problem having much more complicated operators than the 1D case, and a much larger input size $32 \times 32$.

However, for the 2D case the architecture hyperparameters have to be changed so that they can adequately capture the more complicated 2D fields. Hence, for the DeepSets encoder, we use an MLP $\phi$ with a first layer of 1500 neurons, and 3 layers of 1200 neurons each. The pooled latent space has dimension 1200 with mean pooling. The branch network has 2 layers of 1000 neurons, and 3 layers of 800 neurons, and the trunk network has 5 layers of 1000 neurons, and both embed into a dimension 1200 space. All networks use ReLU activation functions. 
Once trained, we evaluate the model by calculating the relative L2 error for 100 randomly sampled operators form the test dataset, and repeat this for various in context sizes to see how the error changes. Comments on performance are found in $\ref{sec:experimental}$

\subsection{Multiple differential equation operators}

We set $u(0) = 0$ for the forward version of the three initial value problems (ODEConst, ODE1, and ODE2). For the boundary value problems (Poisson, Linear Reaction-Diffusion, and Nonlinear Reaction-Diffusion), the boundary values $u(0)$ and $u(1)$ are unknown parameters in both the forward and inverse versions, and must be learned in-context.
% For example, the Poisson boundary-value problem is:
% \begin{equation}
%     \begin{split}
%         \frac{d^2u(x)}{dx^2} &= f(x), \ x \in (0,1) \\
%         &u(0) = u_0 \\
%         &u(1) = u_1
%     \end{split}
% \end{equation}
% %are the boundary conditions
% The forward problem is to find the solution $u(x)$ given the forcing function $f(x)$. The inverse or control problem is to find the forcing function $f(x)$ that produces the given trajectory $u(x)$. Very importantly, the boundary values $u_0$ and $u_1$ are unknown parameters and must be learned in-context.
In our experiments, all parameters (i.e., equation coefficients and boundary values) are chosen randomly. The physical domain in all cases is the interval $[0,1]$, which is discretized uniformly into 100 points. The prompts during both training and testing contain 4 example pairs of parameter and solution functions. The MLP $\phi$ in the DeepOSets contains 2 hidden layers of 500 neurons each, and the latent vector encoding the prompt data has size 500. The pooling operation is the sample mean. The branch and trunk networks in the DeepOSets contain hidden layers of 200, 500, and 200 neurons with intermediate layer of 500 neurons. The pooled latent space is 2000 with mean pooling. All neural networks use the ReLU activation function. 

To generate the training data for each of the problems, 100 parameter functions are sampled independently from a Gaussian process with the squared-exponential kernel (discretized on the uniform 100-point grid) with variance 2 and length scale $0.5$. For each boundary-value problem, the boundary values $u_0$ and $u_1$ are sampled uniformly in the interval $[-1,1]$. For the reaction-diffusion problems, the coefficients $a$ and $c$ are sampled uniformly in the intervals $[0.025, 0.075]$ and $[-2,2]$, respectively. Then the equation is solved for the given parameters to obtain the ground truth solutions.   
The model was trained by minimizing the mean-squared error loss between predicted and ground truth solutions over a mini-batch of 100 operators sampled randomly from the training data at each iteration, using the Adam optimizer with learning rate 1e-3, for 50k iterations. 

To evaluate how a broader task spectrum influences performance, we trained our model on configurations of 6 and 12 tasks. 
Training took 12min for the six-task experiment, achieving a training loss 8.4e-3, while the training time was 38 minutes for the 12-task experiment, achieving a training loss of 4.8e-3. Both experiments were run on an Nvidia 4060 Ti GPU with 16GB memory. We contrast this to the training time required by the complex architecture in \citet{yang2023icon}, which is measured in tens of hours on the same computing hardware.
Sample results are shown in Figure~\ref{fig:1d} and MSE values on test data from operators not seen during training are displayed in Table~\ref{tbl:probs}. DeepOSets achieves accurate predictions without weight updates at inference time, without knowing which task or the values of the equation coefficients and boundary conditions. All these unknowns are correctly disambiguated from the examples in the prompt.

\subsection{ICON comparison}\label{app:icon-details}

 Both DeepOSets and ICON were implemented in PyTorch; the ICON implementation is based on the code released by the ICON authors (a 6-layer Transformer encoder with model dimension 256 and 8 attention heads, which processes the context pairs and the query serialized into a sequence of 850 tokens for $m=5$ context pairs). The DeepOSets model uses 4-hidden-layer ReLU MLPs of width 256 for the set encoder, branch, and trunk networks. Both models are trained with Adam (learning rate $10^{-4}$) on the mean-squared error for the same number of steps, and evaluated on 100 random test samples for $n_{\mathrm{ctx}}\in\{1,\dots,5\}$ in-context examples
 
%\paragraph{Dataset.} 
The dataset 
for the benchmark
%of Appendix \ref{subsec:icon-comparison} 
is generated synthetically using finite differences and Gaussian process priors, with the settings in Table~\ref{tbl:icon-data}. The diffusivity is sampled as $k = \exp(\xi)$ with $\xi\sim\mathcal{GP}(0, k_{\mathrm{RBF}})$, $\sigma = 0.5$, $\ell = 0.4$, on the full grid of $[0,1]$ (52 points including the boundaries). The source term $f(x)$ is sampled from $\mathcal{GP}(0, k_{\mathrm{RBF}})$ with $\sigma = 2.0$, $\ell = 0.4$ on the 50 interior points. The solution $u(x)$ is computed for each $(k,f)$ pair with the Thomas algorithm for the tridiagonal finite-difference system.

\begin{table}[htp]
    \centering
    \small
    \begin{tabular}{lr}
    \toprule
    \textbf{Setting} & \textbf{Value} \\
    \midrule
    Grid points $N$ & 50 (interior) \\
    Total operators & 5,000 \\
    Pairs per operator & 40 \\
    Train / test split (operator level) & 90\% / 10\% \\
    Random seed & 42 \\
    \bottomrule
    \end{tabular}
    \vspace{1ex}
    \caption{Dataset settings for the 1D Darcy-flow benchmark used in the ICON comparison.}
    \label{tbl:icon-data}
\end{table}

The DeepOSets architecture used in the experiment was as follows. 
For a query $(f_q, x)$ and context pairs $(f_i, u_i)_{i=1}^m$, the forward pass is
\begin{equation*}
    h = \frac{1}{m}\sum_{i=1}^m \phi([f_i; u_i]), \qquad \hat{u}(x) = \mathrm{branch}([f_q; h]) \cdot \mathrm{trunk}(x) + b.
\end{equation*}
All sub-networks are 4-hidden-layer MLPs with ReLU activations and 256 neurons per hidden layer: the set encoder $\phi$ maps $\mathbb{R}^{2N}=\mathbb{R}^{100}$ to $\mathbb{R}^{256}$, the branch network maps $\mathbb{R}^{N+256}=\mathbb{R}^{306}$ to $\mathbb{R}^{256}$, and the trunk network maps $\mathbb{R}$ to $\mathbb{R}^{256}$. The model has 894,465 trainable parameters.

%\paragraph{ICON architecture.} 
The ICON architecture used in the experiment is described next.
Context pairs and the query are serialized into a token sequence and processed by a Transformer encoder; the output is read from the final $N$ tokens. For $m$ context pairs with $N$ grid points each, the total sequence length is $(3m+2)\times N = 17\times 50 = 850$ tokens at $m=5$. Each token is a 2D vector $(x_j, v_j)$ projected to the model dimension via a linear layer, and a learned positional embedding (indexed by role: $f$, $u$, placeholder, or query) is added. The hyperparameters are listed in Table~\ref{tbl:icon-arch}; the model has 4,744,193 trainable parameters.

\begin{table}[htp]
    \centering
    \small
    \begin{tabular}{lr}
    \toprule
    \textbf{Hyperparameter} & \textbf{Value} \\
    \midrule
    Model dimension & 256 \\
    Transformer layers & 6 \\
    Attention heads & 8 \\
    FFN widening factor & 4 \\
    Activation & GELU \\
    Normalization & Pre-norm (LayerNorm) \\
    Sequence length & 850 \\
    \bottomrule
    \end{tabular}
    \vspace{1ex}
    \caption{ICON hyperparameters used in the comparison.}
    \label{tbl:icon-arch}
\end{table}

%\paragraph{Training and evaluation protocol.} 
Both models were trained in isolated subprocesses to ensure fair GPU memory separation using Adam with learning rate $10^{-4}$ and the mean-squared error loss, with batch size 5 for DeepOSets and 32 for ICON. At each training step, a random operator index and a random permutation of its pairs are drawn; the first $n_{\mathrm{ctx}}$ pairs form the context and one additional pair is the query. Two equal-step schedules were run, of 10,000 and 100,000 training steps. Evaluation is performed over 100 random test samples for each $n_{\mathrm{ctx}}\in\{1,2,3,4,5\}$, using the relative $\ell^2$ error on the output grid. Each inference forward pass is timed individually with \texttt{torch.cuda.synchronize()} before and after, to obtain accurate GPU wall-clock time; the reported value is the mean over all 500 timed calls (100 per $n_{\mathrm{ctx}}$ level). Table~\ref{tbl:icon-10k} reports the cost measurements of the 10k-step run, which complement the 100k-step measurements of Table~\ref{tbl:icon-cost}.

\begin{table}[htp]
    \centering
    \small
    \begin{tabular}{lrr}
    \toprule
    \textbf{Metric} & \textbf{DeepOSets} & \textbf{ICON} \\
    \midrule
    Trainable parameters & 894,465 & 4,744,193 \\
    Training steps completed & 10,000 & 10,000 \\
    Training time (s) & 42.5 & 7,746.2 \\
    Inference time / sample (ms) & 0.670 & 8.042 \\
    \bottomrule
    \end{tabular}
    \vspace{1ex}
    \caption{Cost measurements for the 10k-step run of the ICON comparison.}
    \label{tbl:icon-10k}
\end{table}

\section{Prompt ablation}\label{subsec:ablations}

To verify that DeepOSets actually uses the in-context examples to disambiguate the operator, we performed prompt ablations on the trained Burgers', 1D Darcy and 2D Darcy models.
\emph{No context:} replacing the pooled prompt embedding by a zero vector causes performance to degrade sharply; the predictions collapse to (approximately) the spatial mean of all dataset solutions, yielding high relative $L_2$ errors in the range $0.45$--$0.62$.
\emph{Mismatched context:} when prompt pairs $(u_i, G_{\mathrm{op}_1}(u_i))$ from one operator are provided while the query function $u_q$ is evaluated under a different operator $G_{\mathrm{op}_2}$, the model outputs the solution corresponding to $G_{\mathrm{op}_1}$, confirming that the prompt controls the operator selection.
\emph{Shuffled context:} permuting the order of the context pairs $(u_i, G(u_i))_{i=1}^m$ yields identical predictions, up to numerical noise below $10^{-7}$, as expected from the exact permutation invariance of the DeepSets encoder.
\emph{Mispaired context:} permuting the outputs $G(u_i)$ while keeping the inputs $u_i$ fixed, so that each input is paired with the wrong output, degrades the predictions substantially.

\section{Nearest-neighbor baseline with sample size ablation}

A natural question is whether, given a large training set, the task could be solved by simple lookup rather than by in-context learning. To test this, we implemented a parameter-free nearest-neighbor (NN) baseline on the 1D Darcy flow problem: given the $n_{\mathrm{ctx}}$ in-context (condition, solution) pairs describing a new test operator, the baseline searches the entire training set for the single operator whose example pairs are jointly closest (in both condition and solution) to the given in-context examples, and then predicts the query solution by returning, from that training operator's own data, the solution whose condition is closest to the query condition. 
The nearest-neighbor baseline has no learned parameters and proceeds in two steps. (1) Given the $n_{\mathrm{ctx}}$ in-context (condition, solution) pairs describing a new test operator, search the entire training set for the single operator whose in-context example pairs are jointly closest (matching both condition and solution in each pair) to the given in-context examples. (2) Predict the query solution by looking up, within that nearest training operator's own data, the pair whose condition is closest to the query condition, and returning its solution. 
For the training-set-size ablation, DeepOSets was retrained from scratch on nested training subsets of $K\in\{100,500,1000,2000,4500\}$ operators with a fixed held-out set of 500 test operators, and the NN baseline used the same nested subsets as its lookup pool. We also ran a training-set-size ablation, retraining DeepOSets from scratch on nested training subsets of $K\in\{100, 500, 1000, 2000, 4500\}$ operators (out of 5,000 total, with a fixed held-out set of 500 test operators never seen during training), and evaluating the NN baseline with the same nested subsets as its lookup pool, so that both methods see identical training data at every scale. The results are reported in Table~\ref{tbl:nn-ablation}.

\begin{table}[htp]
    \centering
    \small
    \begin{subtable}[t]{0.48\textwidth}
        \centering
        \begin{tabular}{crr}
        \toprule
        $K$ (training operators) & DeepOSets & NN baseline \\
        \midrule
        100 & 0.081 & 0.557 \\
        500 & 0.071 & 0.572 \\
        1000 & 0.063 & 0.592 \\
        2000 & 0.082 & 0.545 \\
        4500 (full) & 0.082 & 0.607 \\
        \bottomrule
        \end{tabular}
        \caption{Error at $n_{\mathrm{ctx}}=5$ vs.\ training-set size $K$.}
    \end{subtable}
    \hfill
    \begin{subtable}[t]{0.48\textwidth}
        \centering
        \begin{tabular}{crrr}
        \toprule
        $n_{\mathrm{ctx}}$ & DeepOSets & NN baseline & ICON \\
        \midrule
        1 & 0.210 & 0.692 & 0.134 \\
        2 & 0.118 & 0.613 & 0.083 \\
        3 & 0.098 & 0.563 & 0.059 \\
        4 & 0.082 & 0.654 & 0.057 \\
        5 & 0.082 & 0.607 & 0.060 \\
        \bottomrule
        \end{tabular}
        \caption{Error vs.\ $n_{\mathrm{ctx}}$ at full training-set size $K=4500$.}
    \end{subtable}
    \vspace{1ex}
    \caption{Mean relative $L_2$ test error on the 1D Darcy flow benchmark for DeepOSets and the nearest-neighbor (NN) baseline, on a fixed held-out set of 500 test operators. ICON (trained for 100k steps), is included in (b) as an additional point of reference.}
    \label{tbl:nn-ablation}
\end{table}

The NN baseline's error remains high (relative $L_2$ error between $0.55$ and $0.61$) at every training-set size, and does not shrink as $K$ grows from 100 to 4,500 operators; we see no evidence that the lookup becomes systematically easier simply because the pool is larger. In contrast, DeepOSets achieves a relative $L_2$ error of $0.06$--$0.08$ at every training-set size tested, including the most data-scarce setting of $K=100$ operators (2\% of the full training pool). The margin between DeepOSets and the NN baseline is roughly 7--9$\times$ throughout and does not narrow as more training data is added. These results show that the task is not solved by DeepOSets simply because enough training data is available to make a trivial lookup sufficient, and that the in-context performance of DeepOSets does not depend on having a very large training set.

%\paragraph{Implementation details of the nearest-neighbor baseline.} 

%\pagebreak
\section{Additional Experiments}\label{additional-experiments-appendix}

\begin{figure}[h!]
    \begin{subfigure}[h]{0.5\textwidth}
        \includegraphics[width=\linewidth]{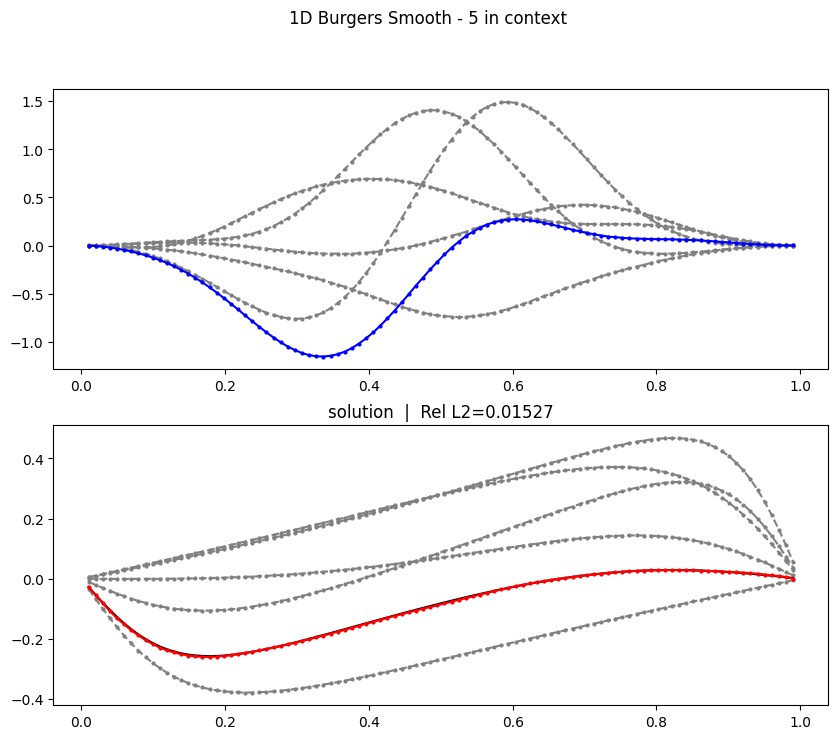}
        \label{fig:burger-smooth}
    \end{subfigure}
    \begin{subfigure}[h]{0.5\textwidth}
        \includegraphics[width=\linewidth]{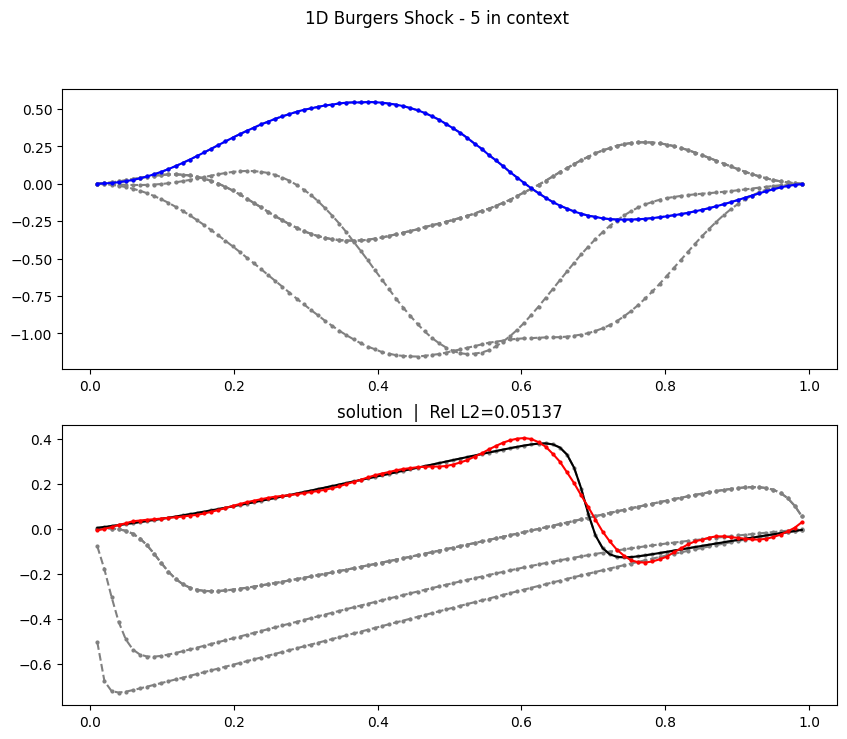}
        \label{fig:burger-shock}
    \end{subfigure}
    \caption{\color{black}Representative testing examples for 1D Burgers, both with 5 in-context examples. Here, the upper graphs with the blue functions represent the in-context query functions, in this case the initial condition $u(x,0)$. The gray functions are the in-context examples $u(x,0)$, and the blue one is the final query. The lower graphs represent the 5 in-context solutions that were given to the network, followed by the network prediction in red, and compared with the true FDM solution in black.} 
    \label{fig:burger}
\end{figure}

\begin{figure}[h!]
    \begin{subfigure}[h]{0.5\textwidth}
        \includegraphics[width=\linewidth]{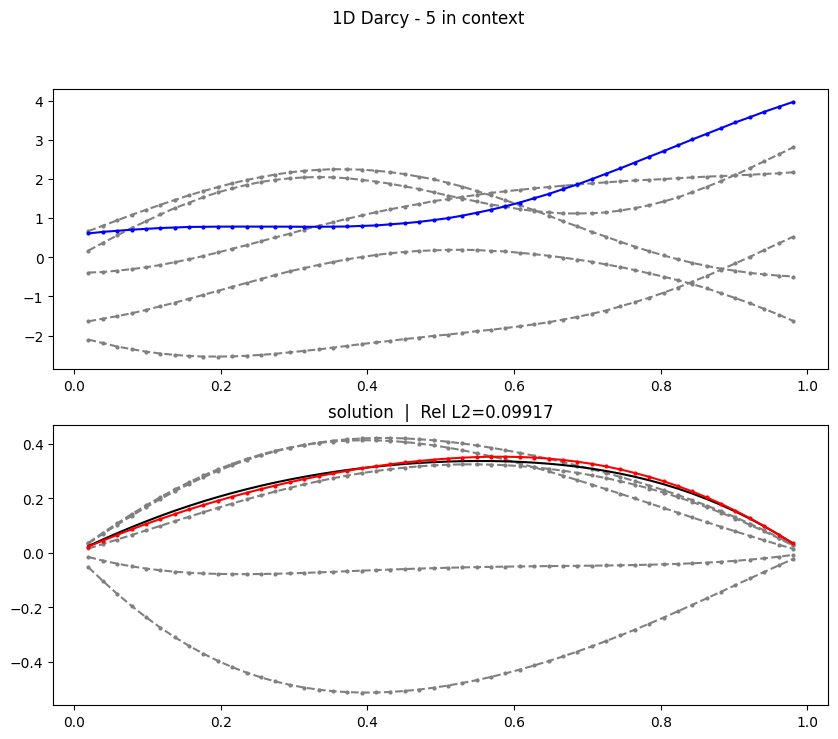}
        \label{fig:darcy-1d-5}
    \end{subfigure}
    \begin{subfigure}[h]{0.5\textwidth}
        \includegraphics[width=\linewidth]{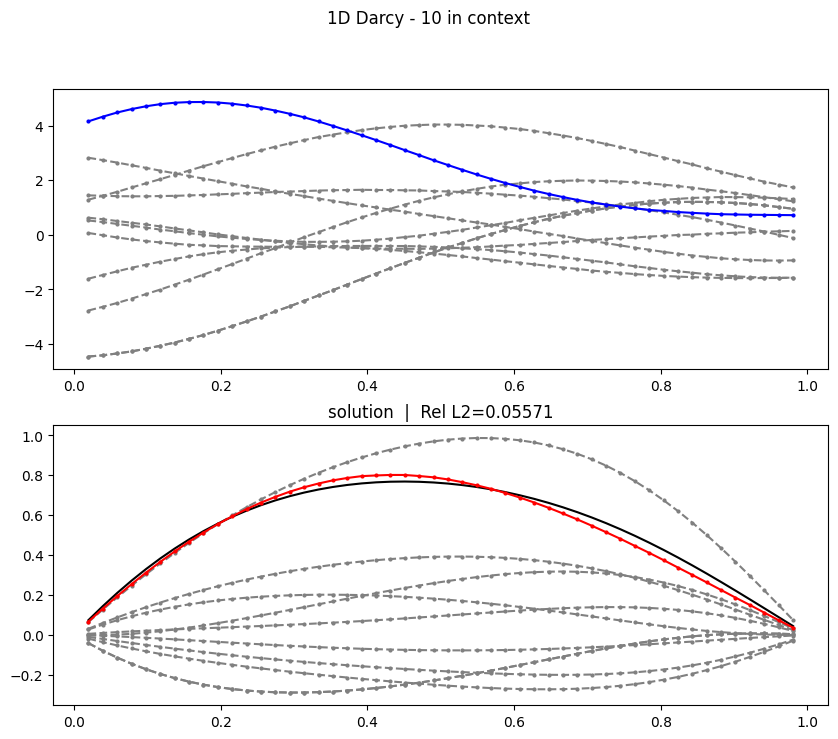}
        \label{fig:darcy-1d-10}
    \end{subfigure}
    \caption{\color{black}Representative testing examples for 1D Darcy with 5 and 10 in-context examples for different operators. The graph is formatted in the exact same manner as the one for Burgers' equation. Here, we see a noticeable reduction in $L_2$ error by simply increasing 5 more examples.} 
    \label{fig:darcy1d}
\end{figure}

\begin{figure}
    \centering
    \includegraphics[width=1\linewidth]{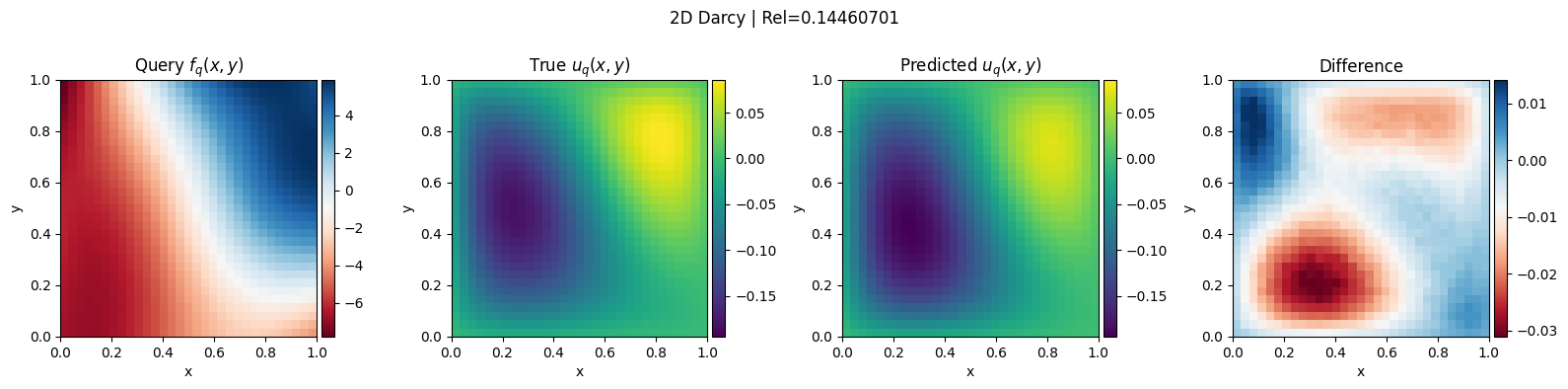}
    \caption{2D Darcy with 5 in-context examples, relative error stated in title, we show the query function, true $u$, our network prediction, and the difference}
    \label{fig:placeholder1}
\end{figure}

\begin{figure}
    \centering
    \includegraphics[width=1\linewidth]{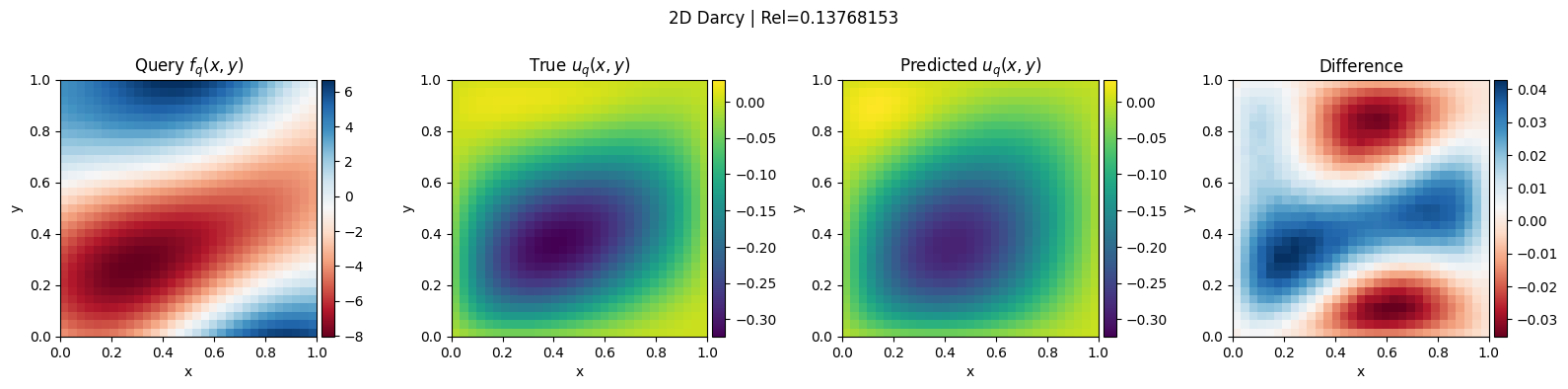}
    \caption{2D Darcy with 10 in-context examples, a slight decrease in relative error, indicating the network still needs additional context}
    \label{fig:placeholder2}
\end{figure}

\begin{figure}
    \centering
    \includegraphics[width=1\linewidth]{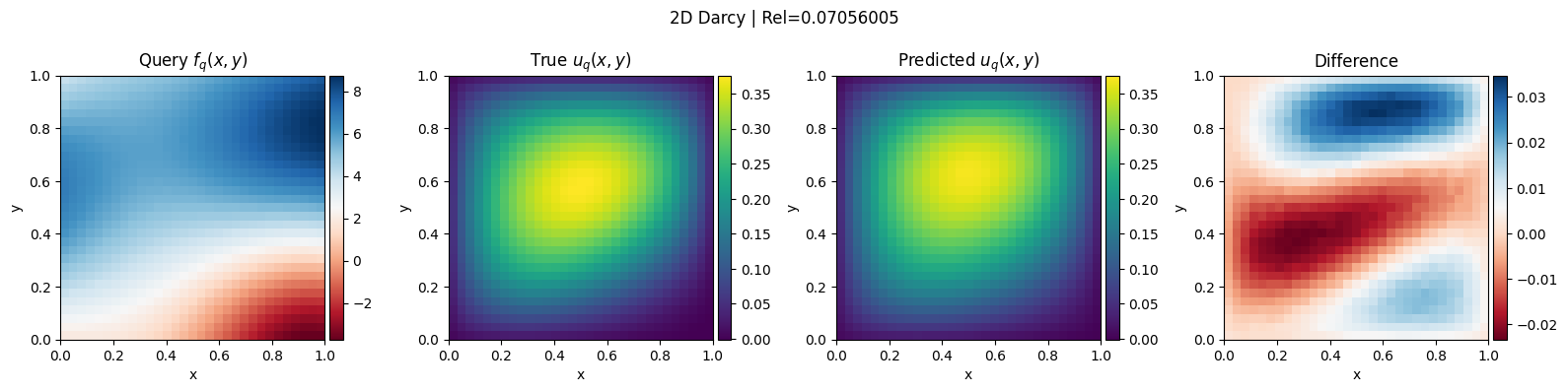}
    \caption{2D Darcy with 20 in-context examples, a much larger decrease in relative error, showing that the network is benefitting from the additional context}
    \label{fig:placeholder3}
\end{figure}

\begin{figure*}[h!]
    \centering
     %\begin{subfigure}[t]{0.4\textwidth}
     %   \caption{Forward ODEConst}
     %   \includegraphics[width=\textwidth]{img_12p_3/pred_odeconst_forward.png}
     %\end{subfigure}
     %\begin{subfigure}[t]{0.4\textwidth}
     %   \caption{Inverse ODEConst}
     %   \includegraphics[width=\textwidth]{img_12p_3/pred_odeconst_inverse.png}
     %\end{subfigure}
    \begin{subfigure}[t]{0.5\textwidth}
        \caption{Forward Poisson}
        \includegraphics[width=\textwidth]{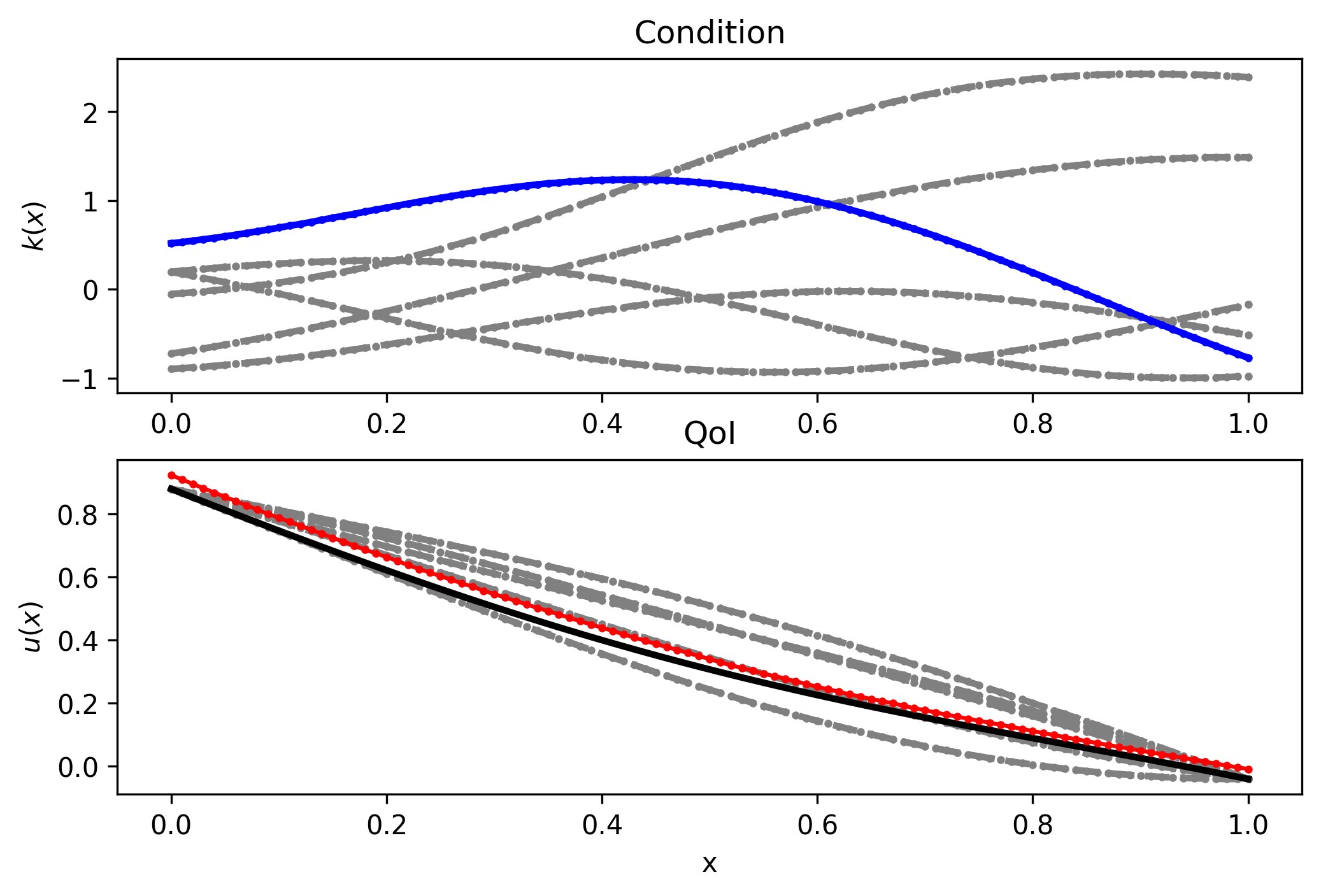}
     \end{subfigure}%
     \begin{subfigure}[t]{0.5\textwidth}
         \caption{Inverse Poisson}
        \includegraphics[width=\textwidth]{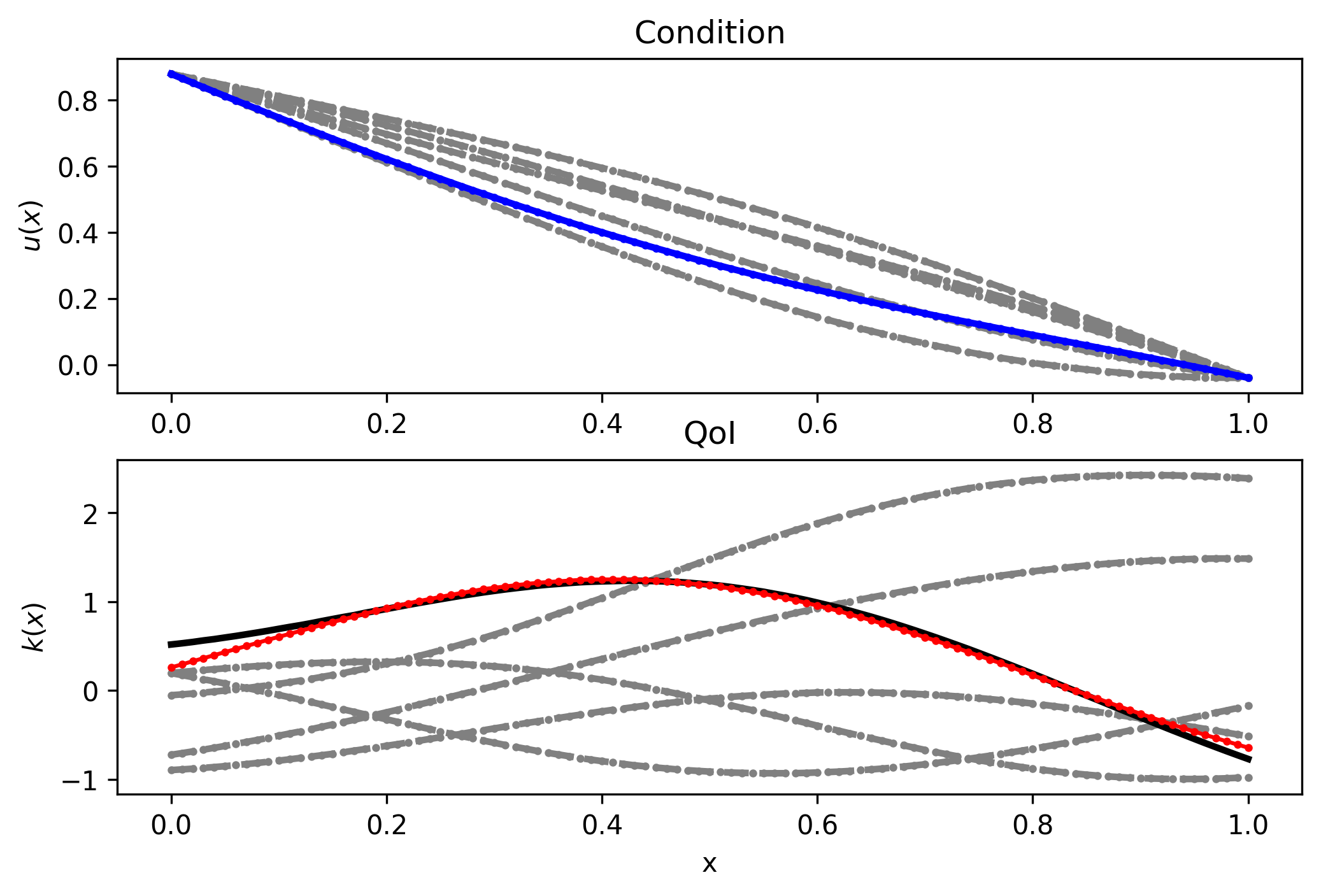}
     \end{subfigure}\\[2ex]
     \begin{subfigure}[t]{0.5\textwidth}
         \caption{Forward reaction-diffusion}
        \includegraphics[width=\textwidth]{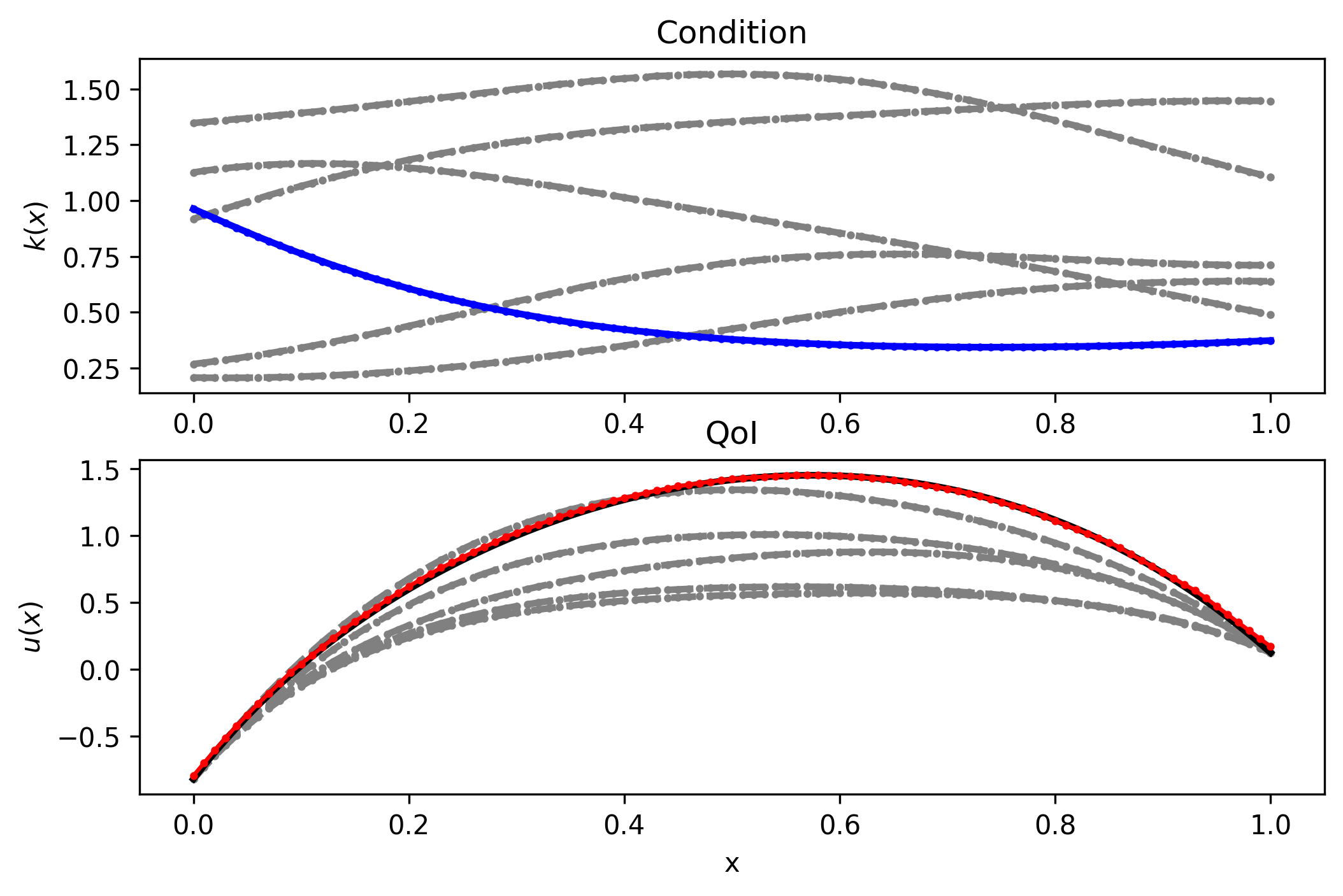}
     \end{subfigure}%
     \begin{subfigure}[t]{0.5\textwidth}
         \caption{Inverse reaction-diffusion}
        \includegraphics[width=\textwidth]{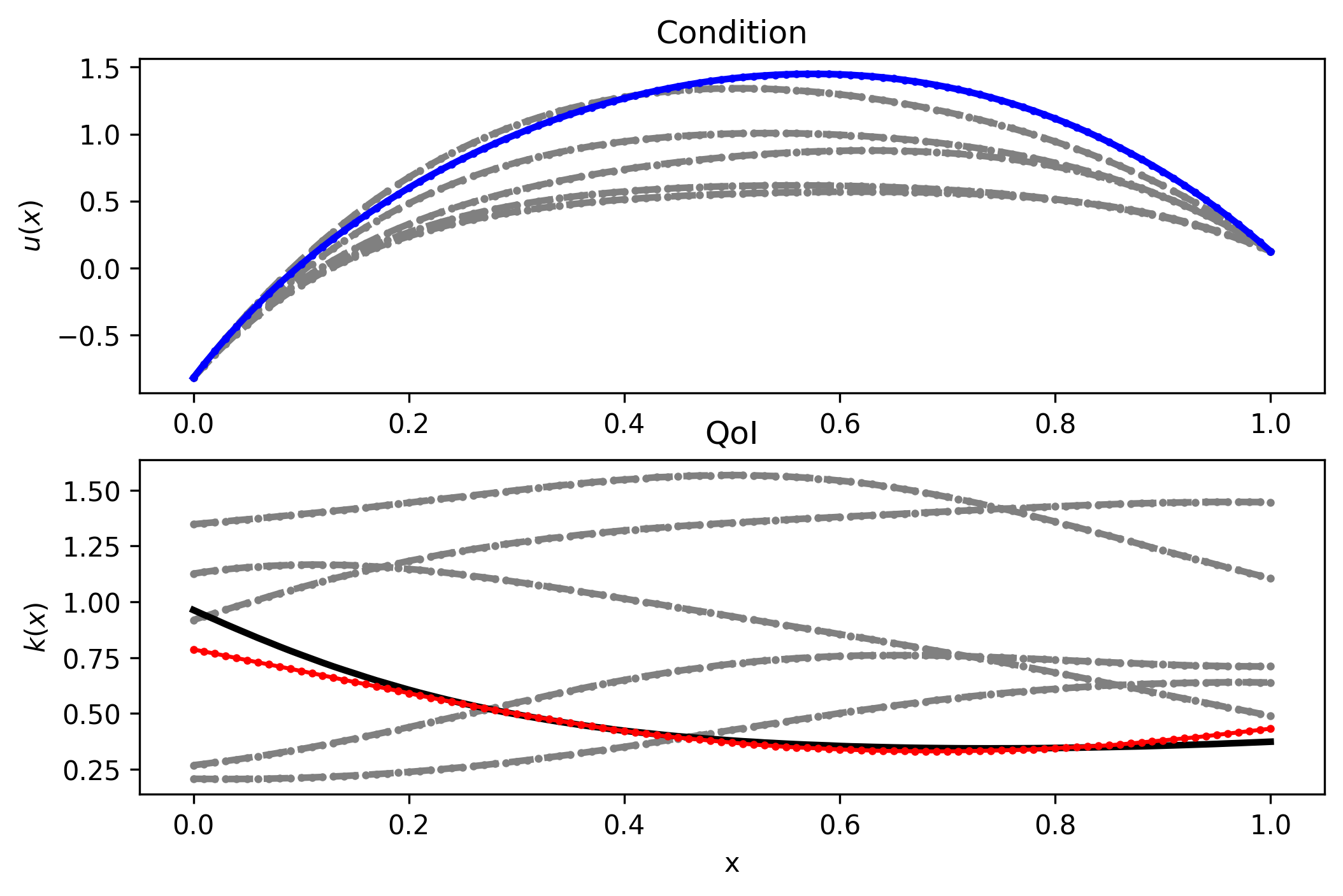}
     \end{subfigure}\\[2ex]
         \begin{subfigure}[t]{0.5\textwidth}
        \caption{Forward nonlinear reaction-diffusion }
        \includegraphics[width=\textwidth]{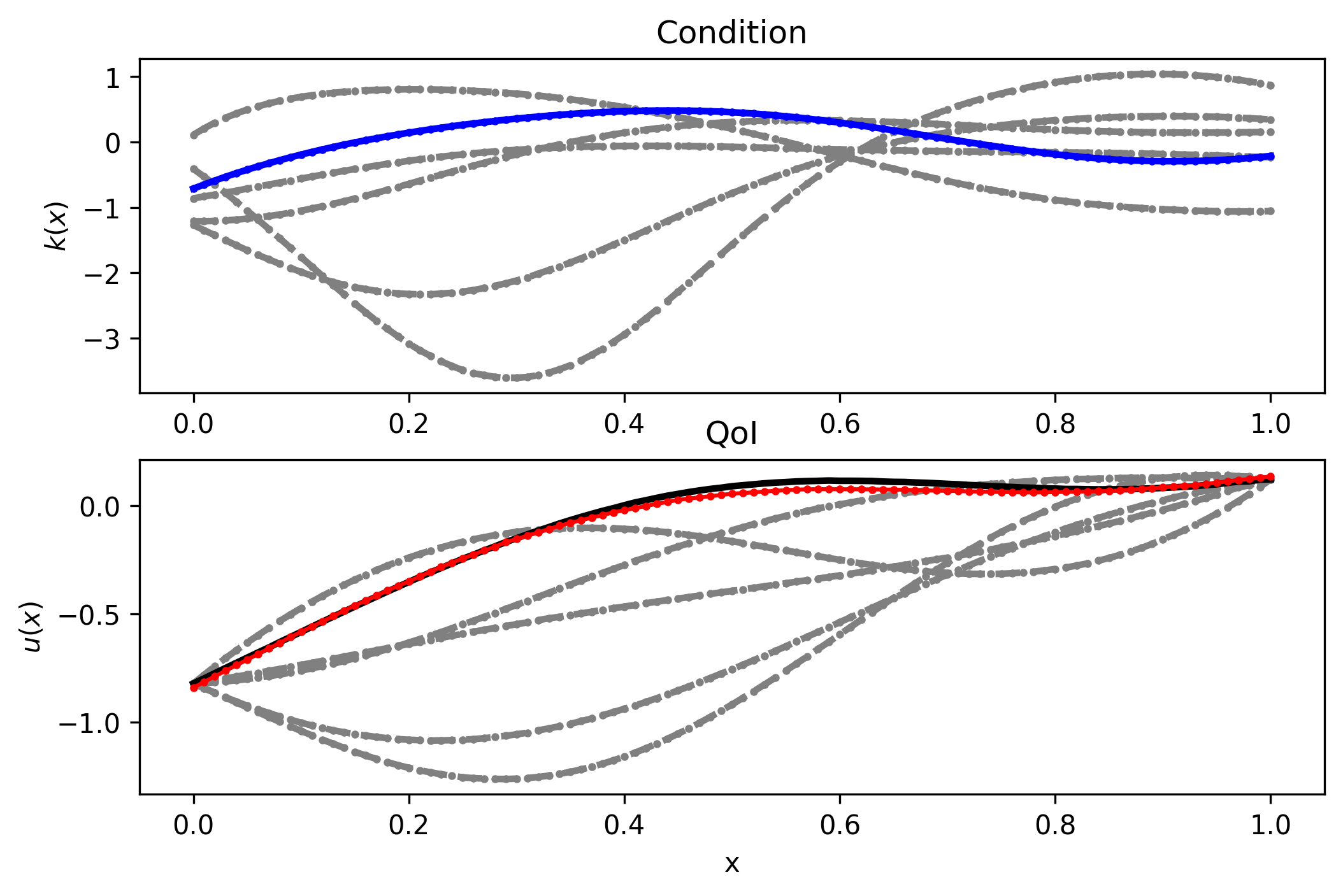}
     \end{subfigure}%
     \begin{subfigure}[t]{0.5\textwidth}
        \caption{Inverse nonlinear reaction-diffusion }
        \includegraphics[width=\textwidth]{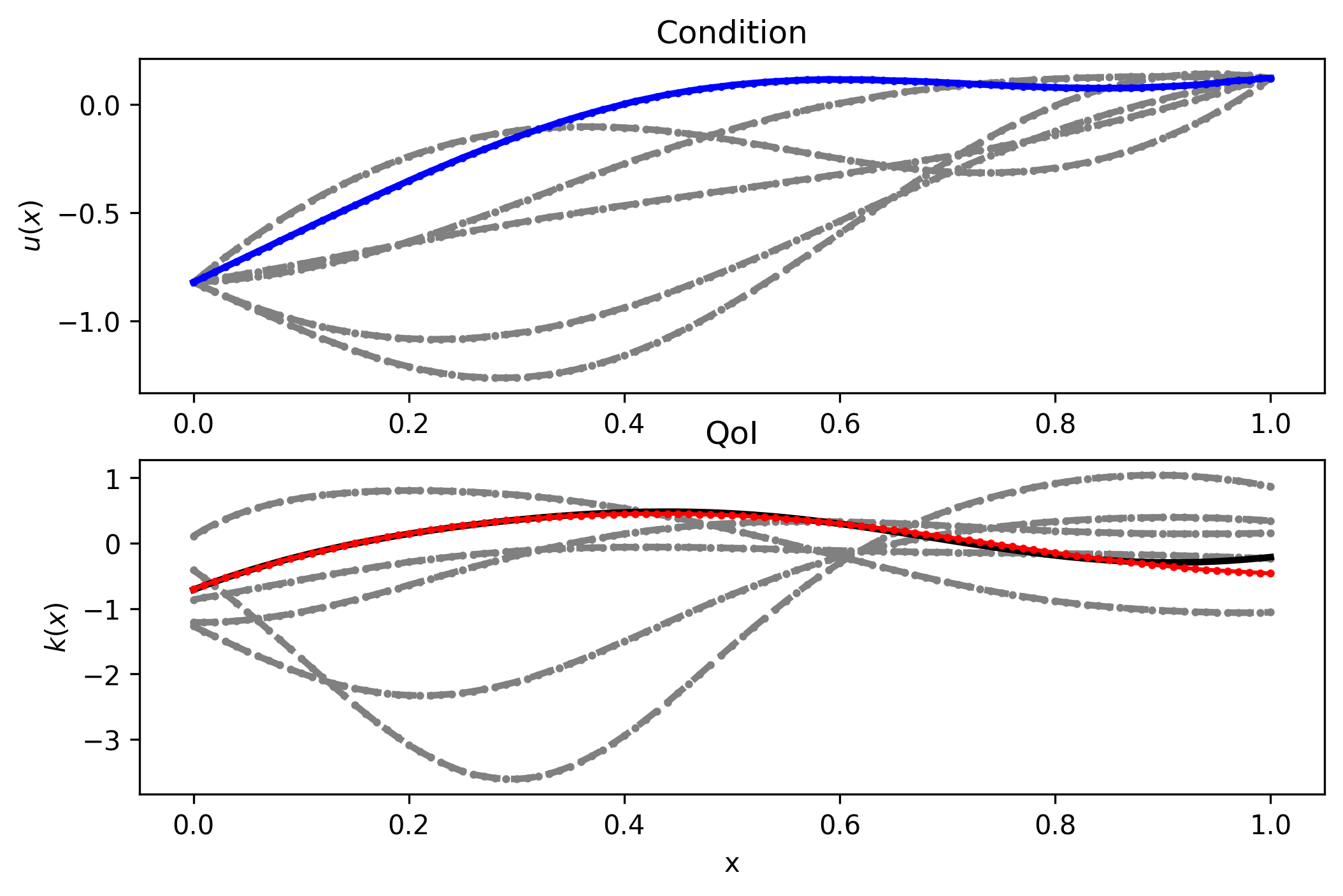}
     \end{subfigure}
     \caption{Sample results from the 12-task numerical experiment. In each case, the top plot contains the parameter functions while the bottom plot contains the corresponding solution functions. The gray lines indicate in each case the parameter-solution function pairs given in the prompt. The blue line indicates the query parameter function, while the red line indicates the corresponding solution predicted by the trained DeepOSets model. The black line indicates the exact solution corresponding to the query, using the appropriate equation.}
    \label{fig:1d}
\end{figure*}

Figures \ref{fig:burger}--\ref{fig:1d} present additional plots and results of the experiments in the paper.

\end{document}